\documentclass[journal]{IEEEtran}
\usepackage{amsmath,amsfonts}
\usepackage{algorithm}
\usepackage{algorithmic}
\usepackage{hyperref}
\usepackage{graphicx}
\usepackage{balance}

\begin{document}

\title{Quaternion Convolutional Neural Networks: Current Advances and Future Directions}
\author{Gerardo Altamirano-Gomez, Carlos Gershenson
\thanks{Instituto de Investigaciones en Matemáticas Aplicadas y Sistemas, UNAM, Mexico}}
\markboth{July~2023}
{Altamirano-Gomez, Gershenson: Quaternion Convolutional Neural Networks: Current Advances and Future Directions}

\maketitle

\begin{abstract}
  Since their first applications, Convolutional Neural Networks (CNNs) have solved problems that have advanced the state-of-the-art in several domains. CNNs represent information using real numbers. Despite encouraging results, theoretical analysis shows that representations such as hyper-complex numbers can achieve richer representational capacities than real numbers, and that Hamilton products can capture intrinsic interchannel relationships. Moreover, in the last few years, experimental research has shown that Quaternion-Valued CNNs (QCNNs) can achieve similar performance with fewer parameters than their real-valued counterparts. This paper condenses research in the development of QCNNs from its very beginnings. We propose a conceptual organization of current trends and analyze the main building blocks used in the design of QCNN models. Based on this conceptual organization, we propose future directions of research.
\end{abstract}


\section{Introduction}
\IEEEPARstart{I}{n} the last decade, the use of deep learning models has become ubiquitous for solving difficult and open problems in science and engineering. Convolutional Neural Networks (CNN) were one of the first deep learning models \cite{goodfellow:2016:deeplearning, khan:2020:cnnsurvey, li:2021:cnnsurvey}, and their success in tackling the large scale object recognition and classification problem (Imagenet challenge) \cite{krizhevsky:2012:alexnet}, led to its application in other domains.

\begin{figure*}[!t]
  \centering
  \includegraphics[width=0.78\textwidth]{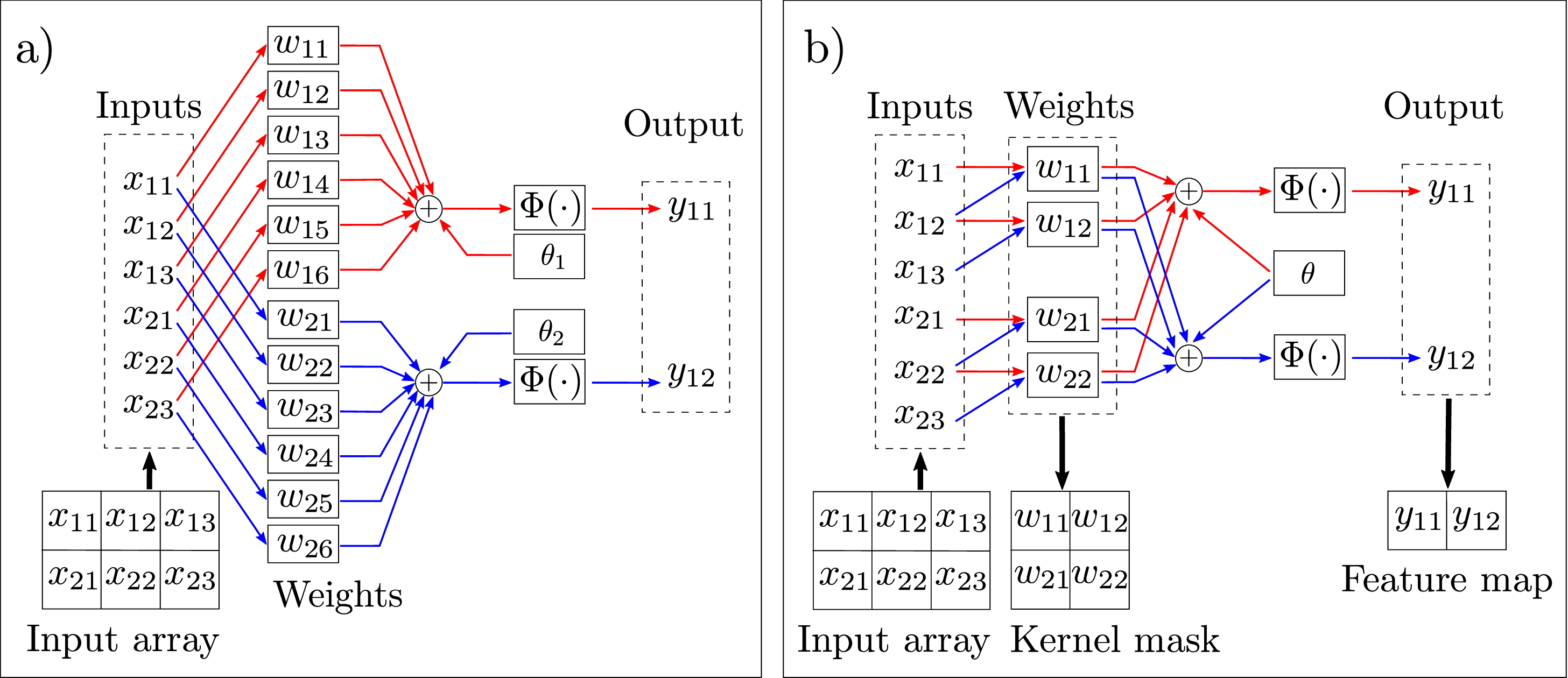}
  \caption {Fully connected layer (a) vs. convolutional layer (b). For an input array of $2\times 3$ elements, the Perceptron uses $12$ weights and $24$ connections (not locally connected,) while the convolution layer uses $4$ weights and $8$ connections. In the convolutional layer, we have a reduction in the number of parameters and connections, because a single weight is connected to several inputs, and the weights are applied over locally connected inputs.}
  \label{fig:perpvsconv}
\end{figure*}

The core components of a CNN architecture are the convolution and pooling layers. A convolution layer is as a variation of a fully connected layer (Perceptron), as shown in Figure \ref{fig:perpvsconv}. In the former case, a weight-sharing mechanism over locally connected inputs is applied \cite{fukushima:1980:neocognitron}. This technique is inspired by the local receptive fields discovered by Hubel and Wiesel in their experiments with macaques \cite{hubelwiesel:1968:receptivefields}.

Formally speaking, the convolution layer applies the mathematical definition of convolution between discrete signals; thus, for a bi-dimensional input:
\begin{equation}
  \mathbf{D}= [d(x,y)] \in \mathbb{R}^{N_1\times N_2},
\end{equation}
the convolution with a kernel:
\begin{equation}
  \mathbf{W}= [w(x,y)] \in \mathbb{R}^{M_1\times M_2},
\end{equation}
is defined as follows:
\begin{eqnarray}
  \mathbf{F} &=& \mathbf{W} \ast \mathbf{D} \nonumber \\
   &=& \sum_{r=-\infty}^\infty \sum_{s=-\infty}^\infty \left[w(r,s) d(x-r,y-s)\right];
\end{eqnarray}
thus, $\mathbf{F}\in\mathbb{R}^{N_1-M_1+1 \times N_2-M_2+1}$ is called \textit{feature map}.

The convolution layer is typically followed by a pooling layer; this provides a sort of local invariance to small rotations and translations of the input features \cite{lecun:1989:designstrategies, lecun:1998:cnn}. Moreover, T. Poggio proved that the combination of convolution and pooling layers produce an invariant signature to a group of geometric transformations \cite{poggio:2012:computationalmagic, poggio:2014:itheoryproves, poggio:2015:itheory}. However, in the design of very deep architectures, researchers have encountered some difficulties, e.g. reducing the number of parameters without losing generalization, and finding fast optimization methods for adjusting millions of parameters avoiding the vanishing and exploding gradient problems \cite{hochreiter:1997:vanishinggradient, pascanu:2013:vanishinggradient}.

Fundamental theoretical, as well as experimental analysis, have shown that some algebraic systems, different from the real numbers, have the potential to solve these problems. For example, using a complex numbers representation avoids local minima caused by the hierarchical structure \cite{nitta:2002:criticalpoints}, exhibits better generalization \cite{hirose:2012:generalization}, and faster learning \cite{arjovsky:2016:complexrnn}. Because of the Cayley-Dickson construction \cite{baez:2002:octonions}, it could be inferred that these properties would hold on quaternion-valued neural networks. Recent experimental work has favored this conjecture, where quaternion-valued neural networks show a reduction in the number of parameters, and improved classification accuracies compared to its real-valued counterparts \cite{arena:1997:qmlp2, gaudet:2018:qcnn, hongo:2020:qcnn, nitta:2004:4bitparity, parcollet:2018:qcnn, shen:2020:qcnn, zhu:2018:qcnn}. In addition, a quaternion representation can deal with four-dimensional signals as a single entity \cite{bulow:2001:quaternionpolar, ell:2014:qftsignalproc, hitzersangwine:2013:qc_fouriertransform, minemoto:2017:qelm, sangwine:2000:hypercomplexfilters}, models efficiently 3D transformations \cite{hanson:2006:quaternions, kuipers:1999:quaternions}, captures internal latent relationships via the Hamilton product \cite{parcollet:2019:qcnn}, among other properties. 

Because of the diversity of deep learning models, this paper focuses in those using quaternion convolution as the main component. Consequently, we have identified three dominant conceptual trends: the classic, the geometric, and the equivariant approaches. These differ in the definition and interpretation of the quaternion convolution layer. The main contributions of this paper can be summarized as follows:

\begin{enumerate}
  \item This paper presents a classification of QCNNs models based on the definition of quaternion convolution.
  \item This paper provides a description of all atomic components needed for implementing QCNNs models, the motivation behind each component, the challenges when they are applied in the quaternion domain, and future directions of research for each component.
  \item This paper presents an organized overview of the models that have been found in the literature. They are organized by application domain, classified in: classic, geometric, or equivariant approach; and presented by the type of model: recurrent, residual, convNet, generative or CAE.
\end{enumerate}

The organization of the paper is as follows: In Section \ref{sec:qalgebra}, we introduce fundamental concepts of quaternion algebra, then in Section \ref{sec:trends} we explain the seminal works on each of the three conceptual trends (classic, geometric, and equivariant), followed by a presentation of the key atomic components to construct QCNNs architectures in Section \ref{sec:qcnnblocks}. Thereafter, in Section \ref{sec:apps}, we show a classification of current works by application, and present the diverse types of architectures. Finally, Section \ref{sec:future} presents open issues and guidelines for future work, followed by the conclusions in Section \ref{sec:conclusions}.

\section{Quaternion algebra}\label{sec:qalgebra}

\begin{figure*}[!t]
  \centering
  \includegraphics[width=0.78\textwidth]{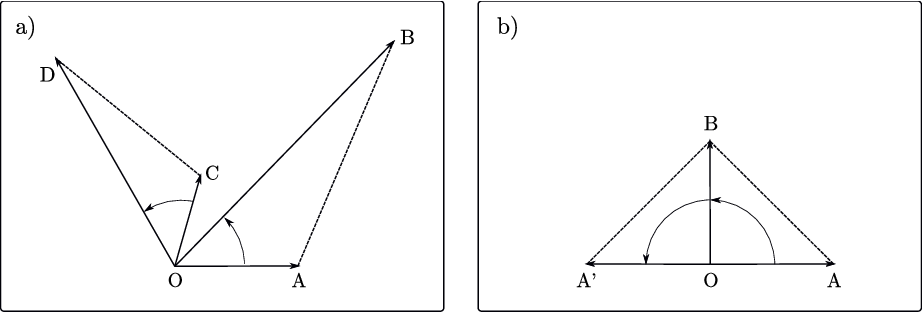}
  \caption {Two similar triangles, turned and in a common plane. a) General configuration b) Right triangles with equal length cathetus configuration. Adapted from \cite{hamilton:1866:quaternions}.}
  \label{fig:elementsp112p130}
\end{figure*}

This mathematical system was developed by W.R. Hamilton (1805-1865) at the middle of the XIX century \cite{hamilton:1853:quaternions, hamilton:1866:quaternions, hamilton:2000:quaternions}. His work on this subject started by exploring ratios between geometric elements, consequently he called \textit{quaternion} to the quotient of two vectors. He also noticed that for two similar triangles lying on a common plane, $AOB$ and $COD$, which are similarly turned, see Figure \ref{fig:elementsp112p130}a), the following equality holds \cite[pp. 112]{hamilton:1866:quaternions}:
\begin{equation}
  \vec{OB}:\vec{OA}=\vec{OD}:\vec{OC}
\end{equation}
or expressed as \textit{geometric fractions}:
\begin{equation}
  \frac{\vec{OB}}{\vec{OA}}=\frac{\vec{OD}}{\vec{OC}},
\end{equation}
where $\vec{OA}$, $\vec{OB}$, $\vec{OC}$, and $\vec{OD}$, are vectors from $O$ to point $A$, $B$, $C$, and $D$, respectively.

Then, by making the triangle $COD$ into $BOA'$, see Figure \ref{fig:elementsp112p130}b); where $\vec{OA}$ and $\vec{OB}$ represents any two equally long vectors, we obtain the following relationship \cite[pp. 130]{hamilton:1866:quaternions}:
\begin{equation}
  \frac{\vec{OB}}{\vec{OA}}= \frac{\vec{OA'}}{\vec{OB}}
\end{equation}
multiplying by $\vec{OB}/\vec{OA}$:
\begin{eqnarray}
  \left(\frac{\vec{OB}}{\vec{OA}}\right)^2 &=& \left(\frac{\vec{OA'}}{\vec{OB}}\right)\left(\frac{\vec{OB}}{\vec{OA}}\right) \nonumber \\
  \left(\frac{\vec{OB}}{\vec{OA}}\right)^2 &=& \frac{\vec{OA'}}{\vec{OA}}
\end{eqnarray}
Since $\vec{OA}$ and $\vec{OA'}$ have the same magnitude, but opposite direction, $\vec{OA}=-\vec{OA'}$, we obtain:
\begin{equation}
  \left(\frac{\vec{OB}}{\vec{OA}}\right)^2=-1.
  \label{eq:qminus1}
\end{equation}

The quotient of two perpendicular equally long vectors, like in this case, is called \textit{right radial quaternion}. Since not particular assumption was made about quotients in Equation (\ref{eq:qminus1}), e.g. the plane of that quotient is arbitrary, Hamilton concluded that every right radial quaternion, was one of the square roots of negative unity \cite[pp. 131]{hamilton:1866:quaternions}. In addition, for a quaternion: $\textbf{q}=\vec{v_1}:\vec{v_2}$, where $\vec{v_1}$ and $\vec{v_2}$ are vectors, we can rewrite it as $\textbf{q} \vec{v_1} = \vec{v_2}$. In this case, $\textbf{q}$ is called a \textit{versor}, i.e. an element that transforms $\vec{v_1}$ into $\vec{v_2}$ by rotating it \cite[pp. 133]{hamilton:1866:quaternions}. In this way, these results connected quaternion representation with the root $\sqrt{-1}$, and its geometric meaning. 

\begin{figure}[!t]
  \centering
  \includegraphics[width=0.35\textwidth]{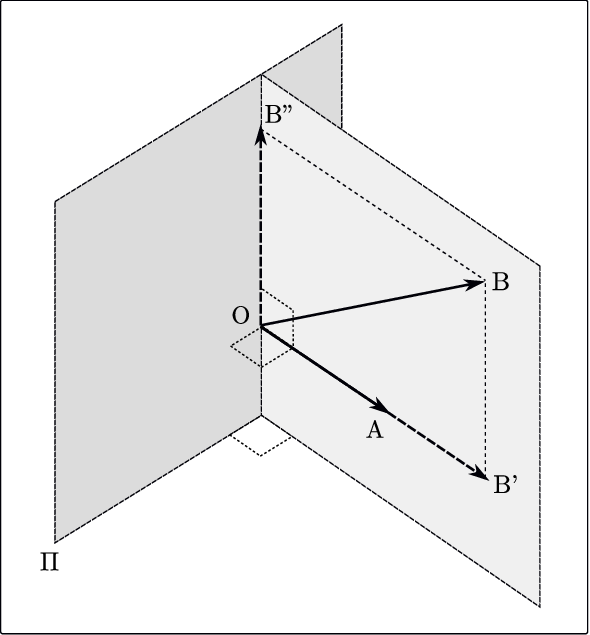}
  \caption {Given two vectors $\vec{OA}$ and $\vec{OB}$, we construct the following geometric configuration: plane $\Pi$ is orthogonal to $\vec{OA}$, vector $\vec{OB''}$ lies on the plane $OAB$ and is the projection of $\vec{OB}$ into $\Pi$, while vector $\vec{OB'}$ is the projection of $\vec{OB}$ into $\vec{OA}$. Adapted from \cite{hamilton:1866:quaternions}.}
  \label{fig:elementsp50}
\end{figure}

Now, considering Figure \ref{fig:elementsp50}, let $\textbf{q}=\vec{OB}:\vec{OA}$ be a quaternion, and let $\vec{OB'}$ and $\vec{OB''}$ be parallel and perpendicular vectors to $\vec{OA}$, respectively; such that $\vec{OB}=\vec{OB'}+\vec{OB''}$. Then, we can decompose the quaternion, $\textbf{q}$, into:
\begin{equation}
  \textbf{q}= \vec{OB'}:\vec{OA} + \vec{OB''}:\vec{OA}
\end{equation}
Since $\vec{OB'}$ and $\vec{OA}$ are parallel vectors, their quotient is just a scale factor representing the projection of $\vec{OA}$ into $\vec{OB'}$, so first term turns into a scalar. The second term represents the projection of $\vec{OA}$ on the plane through $O$, which is perpendicular to $\vec{OA}$; in addition, this means $\vec{OB''}$ can be obtained from $\vec{OA}$ by applying a versor transformation. Since $\vec{OB''}$ and $\vec{OA}$ are perpendicular to each other, its quotient is called \textit{right quaternion} (note that it differs from right radial quaternion in the length of the vectors), and can be expressed as a linear combination of right radial quaternions (versors). This leads to the well know result that every quaternion can be reduced to the quadrinomial form \cite[pp. 160]{hamilton:1866:quaternions}:
\begin{equation}
  \textbf{q} = q_R+ q_I\hat{i}+ q_J\hat{j} + q_K\hat{k},
  \label{eq:quaternion}
\end{equation}
where $q_R, q_I, q_J, q_K$ are scalars, and $\hat{i}, \hat{j}, \hat{k}$ are three right versors.

In terms of modern mathematics, the quaternion algebra, $\mathbb{H}$, is: the 4-dimensional vector space over the field of the real numbers, generated by the basis $\{1,\hat{i},\hat{j},\hat{k}\}$, and endowed with the following multiplication rules (Hamilton product):
\begin{alignat}{6}
  (1)(1)&=& 1 & & & & \nonumber \\ 
  (1)(\hat{i})&=& \hat{j}\hat{k} &=& -\hat{k}\hat{j} &=& \hat{i} \nonumber \\
  (1)(\hat{j})&=& \hat{k}\hat{i} &=& -\hat{i}\hat{k} &=& \hat{j} \nonumber \\
  (1)(\hat{k})&=& \hat{i}\hat{j} &=& -\hat{j}\hat{i} &=& \hat{k} \nonumber \\
  \hat{i}^2 &=& \hat{j}^2 &=& \hat{k}^2 &=&-1
\end{alignat}

Therefore, this definition makes the quaternion algebra associative and non-commutative. Thus, for two arbitrary quaternions: $\textbf{p} = p_R + p_I\hat{i} + p_J\hat{j} + \nonumber p_K\hat{k}$ and $\textbf{q} = q_R + q_I\hat{i} + q_J\hat{j} + q_K\hat{k}$, their multiplication is calculated as follows:
\begin{eqnarray}
  \textbf{p}\textbf{q} &=& p_R q_R - p_I q_I - p_J q_J - p_K q_K \nonumber \\
  && + (p_R q_I + p_I q_R + p_J q_K - p_K q_J) \hat{i} \nonumber \\
  && + (p_R q_J - p_I q_K + p_J q_R + p_K q_I) \hat{j} \nonumber \\
  && + (p_R q_K + p_I q_J - p_J q_I + p_K q_R) \hat{k}
\end{eqnarray}

Notice that each coefficient of the resulting quaternion, is composed of real and imaginary parts of the factors $p$ and $q$. In this way, the Hamilton product capture interchannel relationships between both factors. 

Next, there are introduced some useful operations with quaternions.

Let, $\textbf{q} = q_R+ q_I\hat{i}+ q_J\hat{j} + q_K\hat{k}$, be a quaternion, its \textit{conjugate} is defined as:
\begin{equation}
  \bar{\textbf{q}} = q_R - q_I\hat{i} - q_J\hat{j} - q_K\hat{k}
\end{equation}
And its magnitude is computed as follows:
\begin{equation}
  \|\textbf{q}\|= \sqrt{\textbf{q}\bar{\textbf{q}}}
\end{equation}
As well as complex numbers, quaternions can be represented in polar form \cite{kantor:1989:hypercomplex, ward:1997:quaternionscayley}:
\begin{equation}
  \textbf{q}= \|\textbf{q}\| \left[ \cos(\theta) +\sin(\theta) \frac{q_I\hat{i} + q_J\hat{j} + q_K\hat{k}} {\| q_I\hat{i} + q_J\hat{j} + q_K\hat{k}\|} \right],
  \label{eq:polarform}
\end{equation}
where:
\begin{equation}
  \theta= atan \left( \frac{\sqrt{q_I^2 + q_J^2 + q_K^2}}{q_R} \right).
\end{equation}
Even though other polar parametrizations has been proposed, see for example \cite{altmann:1986:quaternionpolar, bulow:1999:quaternionpolar, bulow:2001:quaternionpolar, sangwine:2010:quaternionpolar}, current QCNNs apply the former equation.

As was mention before, a quaternion can represent a geometric transformation. Let $\textbf{w}_\theta$ be a unitary versor expressed in polar form:
\begin{equation}
  \textbf{w}_\theta=  \cos(\theta) +\sin(\theta) (w_I\hat{i} + w_J\hat{j} + w_K\hat{k}),
\end{equation}
and let $\mathbf{q}$ be any quaternion; then, their multiplication:
\begin{equation}
  \textbf{p}=  \textbf{w}_\theta\textbf{q} 
\end{equation}
applies a rotation, with angle $\theta$, along an axis $\textbf{w}=w_I\hat{i}+w_J\hat{j}+w_K \hat{k}$.

Since we have not made any particular assumption on $\textbf{q}$, we are applying a rotation in the four-dimensional space of quaternions. Denoting by $\Pi_1$ to the plane defined by the scalar axis and the vector $w_I\hat{i} + w_J\hat{j} + w_K\hat{k}$, and $\Pi_2$ the plane perpendicular to $w_I\hat{i} + w_J\hat{j} + w_K\hat{k}$; it can be proved that this 4-dimentional rotation is composed of a simultaneous rotation of the elements on plane $\Pi_1$, and of the elements on plane $\Pi_2$ \cite{ward:1997:quaternionscayley}. 

Alternatively, we can split the transformation as a sandwiching product:
\begin{equation}
  \textbf{p}= \textbf{w}_{\frac{\theta}{2}}\textbf{q}\bar{\textbf{w}}_{\frac{\theta}{2}};
\end{equation}
in this case, the angle of each versor, $\textbf{w}_{\theta/2}$, is divided to half. 

From the group theory perspective, the set of unitary versors lies on a 3-Sphere, $\mathbb{S}^3$, embedded in a 4D Euclidean space \cite{hanson:2006:quaternions}; and together with the Hamilton product form a group, which is isomorphic to the 4D rotation group SO(4) \cite{ward:1997:quaternionscayley}, and to the special unitary group SU(2) \cite{duval:1964:quaternions, hanson:2006:quaternions}. Moreover, there exist a two to one homomorphism with the rotation group SO(3) \cite{duval:1964:quaternions, hanson:2006:quaternions}. 

Another operation of interest is quaternion convolution; because the non-commutative property of quaternion multiplication, we have 3 different definitions of discrete quaternion convolution. First, left-side quaternion convolution, defined as follows \cite{ell:1993:qconv, ell:2007:qconv, pei:2001:qft}:
\begin{equation}
  (\mathbf{w} \ast \mathbf{q})(x,y) = \sum_{r=-\frac{L}{2}}^{\frac{L}{2}} \sum_{s=-\frac{L}{2}}^{\frac{L}{2}} [\mathbf{w}(r,s) \mathbf{q}(x-r,y-s)]
  \label{eq:qcnn:lconvolution}
\end{equation}

Second, right-side quaternion convolution, defined as follows \cite{ell:2007:qconv}:
\begin{equation}\label{eq:qcnn:rconvolution}
  (\mathbf{q} \ast \mathbf{w})(x,y) = \sum_{r=-\frac{L}{2}}^{\frac{L}{2}} \sum_{s=-\frac{L}{2}}^{\frac{L}{2}} [\mathbf{q}(x-r,y-s)\mathbf{w}(r,s)]
\end{equation}

In third place, we have two-sided quaternion convolution \cite{ell:2007:qconv, pei:2001:qft, sangwine:1998:qconv2sided}:
\begin{eqnarray}
  (\mathbf{w_{left}} \ast \mathbf{q} \ast \mathbf{w_{right}})(x,y) = \nonumber\\
  \sum_{r=-\frac{L}{2}}^{\frac{L}{2}} \sum_{s=-\frac{L}{2}}^{\frac{L}{2}} [\mathbf{w_{left}}(r,s)\mathbf{q}(x-r,y-s)\mathbf{w_{right}}(r,s)] \label{eq:qcnn:biconvolution}
\end{eqnarray}
where $q, w, w_{left}, w_{right} \in \mathbb{H}$, $\ast$ represents the convolution operator, and is applied quaternion product between $\mathbf{q}$'s and $\mathbf{w}$'s.

Finally, Table~\ref{tab:notation} summarizes the notation that will be used in the rest of this paper.

\begin{table}
  \caption{Definition of mathematical symbols.}
  \label{tab:notation}
  \begin{tabular}{@{}cc@{}}
    \hline
    Symbol & Meaning\\
    \hline
    $\mathbb{R}$& The filed of the real numbers\\
    $\mathbb{H}$& The quaternion algebra\\
        bold lowercase letter, e,g. $\mathbf{q}$ & A quaternion\\
    bold uppercase letter, e,g. $\mathbf{Q}$ & A quaternionic matrix\\
    $q_R$ & The real component of a quaternion $\mathbf{q}$\\
    $q_I, q_J, q_K$ & The imaginary components of a quaternion $\mathbf{q}$\\
    $\hat{i}, \hat{j}, \hat{k}$ & The imaginary bases of $\mathbb{H}$\\
    $\hat{\cdot}$ & An unitary vector\\
    $\vec{\cdot}$ & A vector\\
    $*$ & Convolution sign\\
    \hline
  \end{tabular}
\end{table}

\section{Development of the classic, geometric, and equivariant approaches.}\label{sec:trends}

In the previous section, it was presented the different definitions of quaternion convolution. From this definition, different conceptual approaches can be obtained by setting some restrictions on the quaternions, e.g. using unitary quaternions, or using just the real part of one of the quaternions. In the following paragraphs, we describe the seminal works that led to the development of the three main conceptual treads found in most of the current works on QCNNs. Since these approaches were not developed in an incremental manner, we focus on tracking the seminal ideas of each approach, describing what components were introduced, and specifying the domain of application in which they were tested. Thereafter, in Section \ref{subsec:qconv}, it is presented the formal definition of each approach.

In the \textit{classic approach}, the definition of quaternion convolution is a natural extension of the real and complex convolution; its role is to compute the correlation between input data and kernel patterns, in the quaternion domain. First works refer to Altamirano \cite{altamirano:2017:geometricperception}, whom based on the work on Quaternion-Valued Multilayer Perceptrons (QMLP's) by Arena \textit{et al.} \cite{arena:1995:qmlp0, arena:1996:qmlp1, arena:1997:qmlp2}, defines the main components of a QCNN: quaternion convolution layers, quaternion pooling layers that use the magnitude of the quaternion, quaternion split-RELU activation function, quaternion fully connected layers, and the quaternion back-propagation training method. As a proof of concept, these components were applied to a simple pattern classification problem. In an independent manner, Gaudet and Maida \cite{gaudet:2018:qcnn} arrived to a similar definition of convolution layers, but their work is based on Travelsi's research on Complex-Valued CNNs \cite{trabelsi:2018:complexnn}; in addition, they proposed quaternion bath-normalization and weight initialization algorithms. Their models were tested by classifying images of the CIFAR-10 and CIFAR-100 databases \cite{krizhevsky:2009:cifar}, as well as the KITTI Road Estimation Benchmark \cite{fritsch:2013:kittiroadestimationbenchmark}. At the same time, and also following the work of Travelsi \cite{trabelsi:2018:complexnn}, T. Parcollet implemented quaternion-valued CNNs \cite{parcollet:2018:qcnn}, and LSTM models \cite{parcollet:2019:qcnnrecurrent}; these were applied on a voice recognition task using the TIMIT Dataset \cite{1993:nist:timitdataset}. In addition, they proposed a decoding-encoding model that converts grayscale images from the KODAK PhotoCD Dataset \cite{franzen:2013:kodakdataset} to RGB images.

Thereafter, Yin \textit{et al.} \cite{yin:2019:qcnn} proposed a quaternion attention mechanism and evaluates it on the image classification task using the CIFAR-10 dataset \cite{krizhevsky:2009:cifar}. Moreover, he proposed a model for detection of double compression JPEG images and tested it on the Uncompressed Color Image Database (UCID) dataset \cite{huang:2010:uciddataset}. Their models implement a pooling method that uses the magnitude of the quaternion, quaternion split activation functions, and an alternative bath normalization method, which reduces the computational cost of other methods by using a single variance value instead of a multichannel covariance matrix.

Alternatively, we have the \textit{geometric approach} which was constructed based on the previous work of Matsui \textit{et al.} \cite{matsui:2004:geometricquaternionnn} and Isokawa \textit{et al.} \cite{ isokawa:2009:quaternionnn}. In this case, the quaternion product applies affine transformations over the input features, consequently the quaternion convolution inherits this property. Thus, Zhu \textit{et al.} \cite{zhu:2018:qcnn} defines quaternion convolution based on a geometric transformation that applies fixed-axis rotation and scaling (2DoF); moreover, they apply the same concept for constructing quaternion fully connected layers. Their model was tested on the image classification problem using the CIFAR-10, CIFAR-100 \cite{krizhevsky:2009:cifar}, and Oxford flowers \cite{nilsback:2008:oxfordflowers} datasets. In addition, they tested their models for the noise elimination task using the Oxford flowers dataset \cite{nilsback:2008:oxfordflowers} and a subset of Microsoft COCO dataset \cite{tsung:2014:mscocodataset}. Under the same approach, Hongo \textit{et al.} \cite{hongo:2020:qcnn} define a quaternion convolution layer that applies affine transformations; in addition, they applied quaternion pooling layers using the magnitude of the quaternion, quaternion split-RELU activation function, and the batch-normalization method of \cite{yin:2019:qcnn}. From these concepts, they implemented QCNNs and residual QCNNs for the image classification problem on the CIFAR-10 dataset \cite{krizhevsky:2009:cifar} . In a recent work, Matsumoto \textit{et al.} \cite{matsumoto:2022:fullyrotationalqcnn} note the limited expression ability of working with fixed axes, and propose a model that learns general rotation axis (4DoF). They applied this model in a pixel classification task with PolSAR images obtained from the Japan Aerospace Exploration Agency (JAXA).

Finally, the third approach is based on the concept of rotation \textit{equivariance}, i.e. if an input feature is rotated a specific angle, then the output feature produced by the convolution layer is equivalent to: take the non-rotated input feature, apply the convolution layer, and then apply the rotation to the output feature. In order to satisfy the rotation equivariant property, Shen \textit{et al.} \cite{shen:2020:qcnn} define a convolution layer involving products between quaternion inputs and real-valued kernels. In addition, they define a rotation equivariant RELU layer and a batch normalization algorithm. Their models were applied to 3D point cloud classification on the ModelNet40 \cite{wu:2015:3dshapenetsdataset} and 3D MNIST \cite{kaggke:2016:3dmnist} datasets. They show that these type of networks are robust to rotated input features, and the feature maps produced by inner layers are invariant to permutations of the input data points.

Another work lying in this category is the one by Jing \textit{et al.} \cite{jing:2021:rotationinvariantqcnn}, who uses the middle element of a convolutional window as a pivot, and define a convolution-like operation that produces rotation equivariant features. By taking the magnitude of the quaternion output, the convolutional blocks can be used to construct rotation invariant classifiers.

Most work on Quaternion-valued CNN is based on these, or lies in one of the preceding categories; in the following section, we will give the formal definition of each approach, and will explain the key atomic components used to construct QCNNs.

\section{QCNN components}\label{sec:qcnnblocks}

In this section, we present the main building blocks for implementing quaternion-valued convolutional deep learning architectures. Future directions for each component are specified at the end of each subsection.

\subsection{Quaternion convolution layers}\label{subsec:qconv}

Lets assume a dataset, where each sample has dimension $N\times M\times 4$, i.e. an input can be represented as a 4-channels matrix of real numbers. Then, each sample, $\mathbf{Q}$, is represented as a $N\times M$ matrix where each element is a quaternion:
\begin{equation}
  \mathbf{Q}= [\mathbf{q}(x,y)] \in \mathbb{H}^{N\times M}
\end{equation}
then,  $\mathbf{Q}$ can be decomposed in its real and imaginary components:
\begin{equation}
  \mathbf{Q}= {Q_R}+{Q_I}\hat{i}+{Q_J}\hat{j}+{Q_K}\hat{k}
\end{equation}
where ${Q_R}, {Q_I}, {Q_J}, {Q_K} \in \mathbb{R}^{N\times M}$, and $\hat{i},\hat{j},\hat{k}$ represent the complex basis of the quaternion algebra.

In the same way, a convolution kernel of size $L\times L$ is represented by a quaternion matrix, as follows:
\begin{equation}
  \mathbf{W}= [\mathbf{w}(x,y)] \in \mathbb{H}^{L\times L}
\end{equation}
which can be decomposed as:
\begin{equation}
  \mathbf{W}= {W_R}+{W_I}\hat{i}+{W_J}\hat{j}+{W_K}\hat{k},
\end{equation}
where ${W_R},{W_I}, {W_J}, {W_K} \in \mathbb{R}^{L\times L}$, and $\hat{i},\hat{j},\hat{k}$ represent the basis of the quaternion algebra.

Then, in the \textit{classic approach}, Altamirano \cite{altamirano:2017:geometricperception} and Gaudet and Maida \cite{gaudet:2018:qcnn} define the convolution layer using left-sided convolution:
\begin{equation}\label{eq:qcnn:qconv1}
  \mathbf{F} = \mathbf{W} \ast \mathbf{Q}.
\end{equation}
Thus, $\mathbf{F}\in \mathbb{H}^{(N-L+1)\times (M-L+1)}$ represents the output of the layer, i.e. a quaternion feature map, and each element of the tensor is computed as follows:
\begin{equation}
  \mathbf{f}(x,y) = (\mathbf{w} \ast \mathbf{q})(x,y).
\end{equation}
This approach does not make any particular assumption about quaternions $\mathbf{w}$ and $\mathbf{q}$. Thus, the convolution represents the integral transformation of a quaternion function on a quaternion input signal.

On contrast, in the \textit{geometric approach}, Zhu \textit{et al.} \cite{zhu:2018:qcnn} apply the two-sided convolution definition:
\begin{equation}\label{eq:qcnn:qconv2}
  \mathbf{F} = \mathbf{W} \ast \mathbf{Q} \ast \mathbf{\bar{W}}
\end{equation}
where $\mathbf{F}\in \mathbb{H}^{(N-L+1)\times (M-L+1)}$, and each element of the output is computed as follows:
\begin{equation}
  \mathbf{f}(x,y) =\sum_{r=-\frac{L}{2}}^{\frac{L}{2}} \sum_{s=-\frac{L}{2}}^{\frac{L}{2}} \frac{\mathbf{w}(r,s)\mathbf{q}(x-r,y-s)\bar{\mathbf{w}}(r,s)}{a_{r,s}}.
\end{equation}

In addition, each quaternion $\mathbf{w}$, is represented in its polar form:
\begin{equation}
  \mathbf{w}(r,s)=a_{r,s}\left( \cos\frac{\theta_{r,s}}{2}+\sin\frac{\theta_{r,s}}{2} \vec{u} \right)
\end{equation}
where $\theta_{r,s}\in[-\pi,\pi]$, $a_{r,s}\in\mathbb{R}$, and $\vec{u}$ represents the unitary rotation axis. Thus, quaternion convolution applies rotation and scaling transformations on the quaternion $\mathbf{q}$. 

Similarly, Hongo \textit{et al.} \cite{hongo:2020:qcnn} use the two-sided quaternion convolution definition, but adds a threshold value for each component of the quaternion:
\begin{equation}\label{eq:qcnn:qconv3}
  \mathbf{F}  = \mathbf{W} \ast \mathbf{Q} \ast \mathbf{\bar{W}} + \mathbf{B}.
\end{equation}
where $\mathbf{B} \in \mathbb{H}$.

In both cases, quaternion convolution applies geometric transformations over the input data.

Finally, in the \textit{equivariant approach} Shen \textit{et al.} \cite{shen:2020:qcnn} propose a simplified version; they convolutes the quaternion input, $\mathbf{Q} \in \mathbb{H}$, with a kernel of real numbers, $\mathbf{W} \in \mathbb{R}$:
\begin{eqnarray}
  \mathbf{F}  &=& \mathbf{W} \ast \mathbf{Q} \nonumber \\ 
  &=& \mathbf{W} \ast \mathbf{Q_R} +\mathbf{W} \ast \mathbf{Q_I} \hat{i}+\mathbf{W} \ast \mathbf{Q_J} \hat{j} \nonumber\\
  && + \mathbf{W} \ast \mathbf{Q_K} \hat{k}.
  \label{eq:qcnn:qconv4}
\end{eqnarray}
This version, allows to extract equivariant features, e.i. if an input sample, $\mathbf{Q}$, produce a feature map, $g(\mathbf{Q})$, then the rotated input sample, $R(\mathbf{Q})$, will produce a rotated feature map, $g(\mathbf{Q})$, thus:
\begin{equation}
  g(R(\mathbf{Q}))= R(g(\mathbf{Q})).
\end{equation}

Independently of the approach that we follows, a related problem when we implement QCNNs, is how to deal with multidimensional inputs. In real-valued CNNs, the common way of dealing with them is: defining 2D convolution kernels with the same number of channels as the input data, apply 2D convolution separately for each channel; then, the resulting 2D outputs are summed over all channels to produce a single-channel feature map. A different approach is to apply N-dimensional convolution; in this case, the multidimensional kernel is convoluted over all the channels of the input data.

For quaternion-valued CNNs, current implementations use variations of the first approach; i.e. the input data is divided into 4-channel sub-inputs, thereafter quaternion convolution is computed for each sub-input. Before explaining the details, some notation is introduced.

Let $\mathbf{X} \in \mathbb{R}^{N\times M \times C}$ be an input data, $N$ is the number of rows, $M$ the number of columns, and $C$ is the number of channels, where $C\%4=0$, then $X$ is partitioned as follows:
\begin{equation}
  \mathbf{X}=  [\mathbf{Q_0}, \mathbf{Q_1}, \dots, \mathbf{Q_{(C/4)-1}} ]
\end{equation}
where each $\mathbf{Q_s} \in \mathbb{H}^{N\times M}$, $0<s<(C/4)-1$ is a \textit{quaternion channel}.

Let $\mathbf{V} \in \mathbb{R}^{L\times L \times K}$, with $K\%4=0$, be the convolution kernel, then:
\begin{equation}
  \mathbf{V}=  [\mathbf{W_0}, \mathbf{W_1}, \dots, \mathbf{W_{(K/4)-1}} ]
\end{equation}
where each $\mathbf{W_s} \in \mathbb{H}^{L\times L}$, $0<s<(K/4)-1$ is a quaternion channel.

\begin{figure}[!t]
  \centering
  \includegraphics[width=0.5\textwidth]{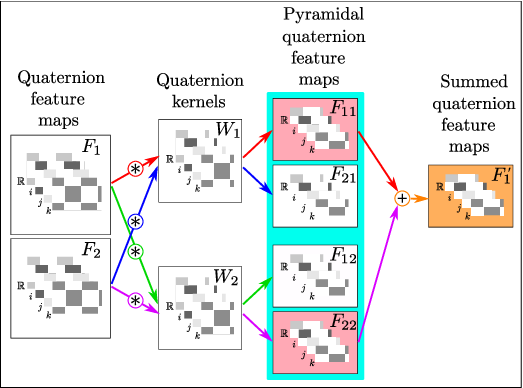}
  \caption {Visualization of different approaches for computing quaternion convolution on a multichannel input. This example shows a 2D input with 2 quaternion channels; the output of an autoencoder convolution layer is shaded in pink, the output of the pyramidal convolution layer is framed in cyan, and the summed convolution layer is shaded in orange.}
  \label{fig:qconvschemes}
\end{figure}

Thus, there are three different ways of dealing with multidimensional quaternion inputs, see Figure \ref{fig:qconvschemes}:
\begin{enumerate}
  \item Autoencoder convolution: Kernel and input have the same number of channels ($K=C$). In this case, each quaternion channel input is assigned to one quaternion channel kernel, and convolution between them is computed using Equation (\ref{eq:qcnn:qconv1}), (\ref{eq:qcnn:qconv2}), (\ref{eq:qcnn:qconv3}) or (\ref{eq:qcnn:qconv4}); Thus, if we use left-sided convolution, each individual output, $\mathbf{F_s} \in \mathbb{H}^{N-L+1\times M-L+1}$, is computed as follows:
  \begin{equation}
    \mathbf{F_s}= \mathbf{W_s} * \mathbf{Q_s},
  \end{equation}
  where $0<i<(C/4)-1$, and the final quaternion feature map is obtained by concatenating all outputs:
  \begin{equation}
    \mathbf{F}=  [\mathbf{F_0}, \mathbf{F_2}, \dots, \mathbf{F_{(C/4)-1}} ]
  \end{equation}
  This method produces an output with the same number of channels as the input data; and could be used in Convolutional Auto-Encoders (CAE).

  \item Pyramidal convolution: Kernel and input have different number of channels ($K\neq C$), but they are multiples of $4$. In \cite{altamirano:2017:geometricperception}, it is proposed the computing of feature maps using a pyramidal approach: each kernel $\mathbf{W_t}$, where $0<t<(K/4)-1$ is convolved with each sub-input $\mathbf{Q_s}$, where $0<s<(C/4)-1$, hence, it produces $C/4$ quaternion outputs. Since each quaternion input channel is convolved with each quaternion kernel channel, see Figure \ref{fig:qconvschemes}, a convolution kernel $\mathbf{V} \in \mathbb{R}^{L\times L \times K}$ will produce $C*K/16$ quaternion outputs:
  \begin{equation}
    \mathbf{F}=  [\mathbf{F_0}, \mathbf{F_2}, \dots, \mathbf{F_{C*K/16}}].
  \end{equation}
  Thus, if left-sided convolution is applied, each quaternion output channel, $\mathbf{F_k}$, is computed as follows:
  \begin{equation}
    \mathbf{F_{(t*C/4)+s}}= \mathbf{W_t} * \mathbf{Q_s}.
  \end{equation}
  Even though \cite{altamirano:2017:geometricperception} used left-sided convolution, this approach is valid with any of the other convolution definitions. The intuition behind this approach is to detect a quaternionic pattern in any sub-input; however, its application is impractical Beacuse of the exponential growth of the number of channels in deep architectures. Calculation of summed outputs can alleviate the computational cost.

  \item Summed convolution: Similarly to the former method, but in this approach, each quaternion input channel is convolved with a different quaternion kernel channel: 
  \begin{equation}
    \mathbf{F_s}= \mathbf{W_s} * \mathbf{Q_s},
  \end{equation}
  where $0<s<(C/4)-1$. Then, the final quaternion feature map, $\mathbf{F}\in \mathbb{H}^{N\times M}$, is obtained by summing all outputs:
  \begin{equation}
    \mathbf{F}  =  \sum_{s=0}^{(C/4)-1} \mathbf{F_s}.
  \end{equation}
\end{enumerate}

\textit{Future directions:} As was mentioned before, N-dimensional convolution has been applied on Real-Valued CNNs for processing multichannel inputs. In a similar way, the equations presented can be extended to 8-channel inputs using an octonion or hypercomplex algebra, see for example \cite{grassucci:2021:hypercomplexcnn, sfikas:2022:hypercomplex, vieira:2022:hypercomplexcnn, zhang:2021:hypercomplexcnn}. For larger number of inputs, a geometric algebra \cite{hestenes:1987:ca2gc} representation can be applied. To the best of our knowledge, this type of architecture has not been published to date, but a first approach in this direction can be found in \cite{grassucci:2021:hypercomplexparam, zhang2021:hypercomplexparam}. In these type of deep learning architectures: quaternion, hyper-complex, or geometric, a major concern is the selection of the signature of the algebra, which will embed data into different geometric spaces, and the processing will take distinct meanings accordingly. Thus, a sensible selection of the dimension and signature should be made according to the nature of the problem and the meaning of input data as well as intermediate layers.

\subsection{Quaternion Fully Connected Layers}

Let $\mathbf{Q}$ be a $N_1\times N_2 \times N_3$ tensor, representing the input to a fully connected layer; then each element of $\mathbf{Q}$ is a quaternion:
\begin{equation}
  \mathbf{Q}= [\mathbf{q}(x,y,z)] \in \mathbb{H}^{N_1\times N_2 \times N_3},
\end{equation}
where $N_1, N_2, N_3$ are the height, width, and number of channels of the input. 

Now, for the fully connected layer, it is defined a quaternion kernel, $\mathbf{W}$, of size $N_1\times N_2 \times N_3$, where each element is a quaternion:
\begin{equation}
  \mathbf{W}= [\mathbf{w}(x,y,z)] \in \mathbb{H}^{N_1\times N_2 \times N_3}.
\end{equation}
where $N_1, N_2, N_3$ are the height, width, and number of channels of the input. 

Note that elements of the input and weight tensors are denoted as $q(x,y,z)$, and $w(x,y,z)$, respectively; and the output of the layer will be a quaternion, $\mathbf{f} \in \mathbb{H}$.

Thus, for classic fully connected layers, the output, $\mathbf{f}$, is computed as follows \cite{altamirano:2017:geometricperception}:
\begin{equation}
  \mathbf{f} = \sum_{r,s,t}^{N_1,N_2,N_3} \left[\mathbf{w}(r,s,t) \mathbf{q}(r,s,t)\right].
\end{equation}

Similarly, for geometric fully connected layers, the output is computed as follows \cite{zhu:2018:qcnn}: 
\begin{equation}
  \mathbf{f} =  \sum_{r,s,t}^{N_1,N_2,N_3} \left[ \frac{1}{\| \mathbf{w}(r,s,t)\|} \mathbf{w}(r,s,t) \mathbf{q}(r,s,t) \bar{\mathbf{w}}(r,s,t) \right].
\end{equation}

The difference between quaternion-valued and real-valued fully connected layers relies on the application of the Hamilton product, which captures interchannel relationships, and in the former case, the output is a quaternion. Moreover, it should be noticed that for real-based networks, fully connected layers are equivalent to inner product layers, but for quaternion-valued networks, the output of quaternion fully connected layers and quaternion inner product layers are different.

\textit{Future directions:} Current implementations of fully quaternion layers follows a classic or geometric approach; the equivariance approach should be implemented and tested.

\subsection{Quaternion Pooling}

Most of the current methods for applying quaternion pooling rely on channel-wise pooling, in a similar way as is applied in Real-Valued CNNs. For example, \cite{gaudet:2018:qcnn, matsumoto:2022:fullyrotationalqcnn} use channel-wise global average pooling layers, and \cite{lin:2013:networkinnetwork, lin:2013:globalpooling, zhu:2018:qcnn} applied channel-wise average as well as channel-wise max pooling layers, while \cite{hongo:2020:qcnn, muppidi:2021:qcnn, parcollet:2018:qcnn} applied just channel-wise max pooling layers, defined as follows:
\begin{eqnarray}
  splitPool(\mathbf{Q}) &=& \max_{(x,y)}(q_R(x,y)) + \max_{(x,y)}(q_I(x,y))\hat{i} \nonumber\\
  && + \max_{(x,y)}(q_J(x,y))\hat{j} \nonumber\\
  &&+ \max_{(x,y)}(q_K(x,y))\hat{k},
\end{eqnarray}
where $\textbf{Q}$ is a quaternion submatrix.

In contrast, \cite{altamirano:2017:geometricperception, shen:2020:qcnn, yin:2019:qcnn} use a max-pooling approach, but instead of using the channel-wise maximum value, they select the quaternion with maximum amplitude within a region:
\begin{eqnarray}
  FullyPool(\mathbf{Q}) &=& \mathbf{q}(\bar{x},\bar{y})\ s.t.\ (\bar{x},\bar{y}) \nonumber \\
  &=& \underset{(x,y)}{\arg\max} (\|\mathbf{q}(x,y)\|).
\end{eqnarray}
In addition, since this method can obtain multiple maximum amplitude quaternions, \cite{yin:2019:qcnn} applies the angle cosine theorem to discriminate between them.

\textit{Future directions:} Future works should introduce novel fully quaternion pooling methods, emphasizing the use of interchannel relationships. For example: using polar representations, introducing quaternion measures \cite{luna:2020:quaternionmeasuretheory}, or taking inspiration from information theory.

\subsection{Quaternion batch normalization}

Internal covariance shift \cite{ioffe:2015:batchnormalization}, is a statistical phenomenon that occurs during training: the change of the network parameters causes changes in the statistical distribution of the inputs in hidden layers. Whitening procedures \cite{lecun:1998:efficientbackprop, wiesler:2011:whiteningtheory}, i.e. apply linear transformations to obtain uncorrelated inputs with zero means and unit variances, alleviate this phenomenon. Since whitening the input layers is computationally expensive, because of the computing of covariance matrices, its inverse square root, and the derivatives of these transformations, Ioffe and Szegedy \cite{ioffe:2015:batchnormalization} introduced the batch normalization algorithm, which normalizes each dimension independently. It uses mini-batchs to estimate the mean and variance of each channel, and transform the channel to have zero mean and unit variance using the following equation:
\begin{equation}
  \tilde{x}=\frac{x-E(x)}{\sqrt{\sigma^2+\epsilon}}
\end{equation}
where $\epsilon$ is a constant added for numerical stability. In order to maintain the representational ability of the network, an affine transformation with two learnable parameters, $\gamma$ and $\beta$, is applied:
\begin{equation}
  BN(\tilde{x})=\gamma \tilde{x}+\beta
\end{equation}
Even though this method does not produce uncorrelated inputs, it improves convergence time by enabling higher learning rates, and has allowed the training of deeper neural network models. On the other hand, uncorrelated inputs reduce overfitting and improve generalization \cite{cogswell:2016:decorrelation_cnn}.

Since channel-wise normalization does not assure equal variance in the real and imaginary components, Gaudet and Maida \cite{gaudet:2018:qcnn} proposed a quaternion batch-normalization algorithm using the whitening approach \cite{kessy:2018:whitening}, and treat each component of the quaternion as an element of a four dimensional vector. We call this approach \textit{Whitening Quaternion Batch-Normalization (WQBN)}. Let $\mathbf{x}$ be a quaternion input variable, $\mathbf{x}=[x_R, x_I, x_J, x_K]^T$, $E(\mathbf{x})$ its expected value, and $V(\mathbf{x})$ its quaternion covariance matrix, both computed over a mini-batch, then:
\begin{equation}\label{eq:qcovmatrix}
  V(\mathbf{x}) = 
  \begin{bmatrix}
    v_{rr} & v_{ri} & v_{rj} & v_{rk}\\
    v_{ir} & v_{ii} & v_{ij} & v_{ik}\\
    v_{jr} & v_{ji} & v_{jj} & v_{jk}\\
    v_{kr} & v_{ki} & v_{kj} & v_{kk}
  \end{bmatrix},
\end{equation}
where subscripts represent the covariance between real or imaginary components of $\mathbf{x}$, e.g. $v_{ij}=cov(x_I,x_J)$. Thus, the Cholesky decomposition of $V^{-1}$  is computed, and one of the resulting matrices, $W$, is selected. Thereafter, the whitened quaternion variable, $\tilde{\mathbf{x}}$, is calculated with the following matrix multiplication \cite{gaudet:2018:qcnn}:
\begin{equation}
  \tilde{\mathbf{x}}=W(\mathbf{x}-E(\mathbf{x})),
\end{equation}
 and finally:
\begin{equation}
  WQBN(\tilde{\mathbf{x}})=\mathbf{\Gamma} \tilde{\mathbf{x}}+\mathbf{\beta}
\end{equation}
where $\beta\in\mathbb{H}$ is a trainable parameter, and:
\begin{equation}
  \mathbf{\Gamma} = 
  \begin{bmatrix}
    \Gamma_{rr} & \Gamma_{ri} & \Gamma_{rj} & \Gamma_{rk}\\
    \Gamma_{ri} & \Gamma_{ii} & \Gamma_{ij} & \Gamma_{ik}\\
    \Gamma_{rj} & \Gamma_{ij} & \Gamma_{jj} & \Gamma_{jk}\\
    \Gamma_{rk} & \Gamma_{ik} & \Gamma_{jk} & \Gamma_{kk}
  \end{bmatrix},
\end{equation}
is a symmetric matrix with trainable parameters.

In constrast, Yin \textit{et al.} \cite{yin:2019:qcnn} applied the quaternion variance definition proposed by \cite{wang:2019:quaternioncovariance}:
\begin{equation}\label{eq:qbnyin1}
  V(\mathbf{x}) = \frac{1}{T}\sum_{i=1}^T \mathbf{v} \bar{\mathbf{v}},
\end{equation}
where $\mathbf{v}=\mathbf{x}-E(\mathbf{x})$. Note that in this case, the variance is a single real value. We call this approach the \textit{Variance Quaternion Batch Normalization (VQBN)}. Thus, the batch normalization is computed as follows:
\begin{equation}\label{eq:qbnyin2}
  VQBN(\mathbf{x})=\mathbf{\gamma} \frac{\mathbf{x}-E(\mathbf{x})}{\sqrt{V(\mathbf{x})^2+\epsilon}}+\mathbf{\beta},
\end{equation}
where $\mathbf{\gamma,\epsilon}\in\mathbb{R}$, $\mathbf{\beta}\in\mathbb{H}$ are trainable parameters.

Recently, Grassucci \textit{et al.} \cite{grassucci:2022:qsngan} noted that for proper quaternion random variables \cite{cheong:2011:qproper, via:2010:qproper}, the covariance matrix in Equation (\ref{eq:qcovmatrix}) becomes the diagonal matrix:
\begin{equation}
  V(\mathbf{x}) =4\sigma^2 I,
\end{equation}
and the batch-normalization procedure is simplified to applying Equations (\ref{eq:qbnyin1}) and (\ref{eq:qbnyin2}) with $V(\mathbf{x})=2\sigma$.

A third approach is the one proposed by \cite{shen:2020:qcnn}, who define the batch-normalization operation to be rotation-equivariance, thus:
\begin{equation}
  RQBN(\mathbf{x})=\frac{\mathbf{x}}{\sqrt{E(\|\mathbf{x}\|^2)+\epsilon}}.
\end{equation}

Generative Adversarial Networks apply another type of normalization, called Spectral Normalization \cite{miyato:2018:spectralnormalization}. In these networks, having a Lipschitz-bounded discriminative function is crucial to mitigate the gradient explosion problem \cite{arjovsky:2017:wassersteingan, gouk:2021:lipschitzbatchnorm, zhou:2019:lipschitzgan}; thus, based on their real-valued counterpart, Grassucci \textit{et al.} \cite{grassucci:2022:qsngan} proposed a Quaternion Spectral Normalization algorithm that constrain the spectral norm of each layer. To explain this procedure, we introduce some definitions.

Let $f$ be a generic function, it is K-Lipschitz continuous if, for any two points, $x_1,x_2$, it satisfies:

\begin{equation}
  \frac{\| f(x_1)-f(x_2) \| }{\lvert x_1-x_2\rvert}\leq K
\end{equation}

Let $\sigma(\cdot)$ be the spectral norm of a matrix, i.e. the largest singular value of a matrix, the Lipschitz norm of a function $f$, denoted by $\|f\|_{Lip}$, is defined as follows:
\begin{equation}
  \|f\|_{Lip}=\sup_x \sigma(\nabla f(x)).
\end{equation}

Thus, for a generic linear layer, $f(h)=Wx+b$, whose gradient is $W$, their Lipschitz norm is:
\begin{equation}
  \|f\|_{Lip}= \sigma(W).
\end{equation}

Now, let $\mathbf{W}$ be a quaternion matrix; their Lipschitz norm, $\sigma(\mathbf{W})$, is computed by estimating the largest singular value of $\mathbf{W}$ via the power iteration method \cite{grassucci:2022:qsngan}. Then, Quaternion Spectral Normalization is applied, in a split way, using the following equations:
\begin{eqnarray}
  \bar{W}_R&=&\frac{W_R}{\sigma(W)}, 
  \bar{W}_I=\frac{W_I}{\sigma(W)}, 
  \bar{W}_J=\frac{W_J}{\sigma(W)}, \text{\ and} \nonumber\\
  \bar{W}_K&=&\frac{W_K}{\sigma(W)}.
\end{eqnarray}

For applying the Lipschitz bound to the whole network, it is used the following relationship:
\begin{equation}
  \|f_1\circ f_2\|_{Lip} \leq \|f_1\|_{Lip}\cdot\|f_2\|_{Lip}.
\end{equation}
provided the Lipschitz norm of each layer is constrained to $1$ \cite{grassucci:2022:qsngan}.

\textit{Future directions:} The WQBN algorithm produces uncorrelated, zero mean, and unit variance inputs, but is computationally expensive. In addition, Kessy \textit{et al.} \cite{kessy:2018:whitening} states that there are infinitely many possible matrices satisfying $V^{-1}=\bar{W}W$. In contrast, VQBN algorithm produces zero mean and unit variance input (according to the Wang \textit{et al.} definition \cite{wang:2019:quaternioncovariance}, which averages the variance of the four channels), but since we use a single value variance, scaling is isotropic, and the input channels still correlated. However, this approach greatly reduces the computational time, by avoiding the decomposition of the covariance matrix. Further theoretical and experimental analysis is required to grasp its advantages versus independent channel batch-normalization.

\subsection{Activation functions}

Biological neurons produce an output signal if a set of input stimuli surpasses a threshold value within a lapse of time; for artificial neurons, the role of the activation function is to simulate this triggering action. Mathematically, this behavior is modeled by a mapping: in the case of real-valued neurons, the domain and image is the field of real numbers, while complex and quaternion neurons map complex or quaternion inputs to complex or quaternion outputs, respectively.

Since back-propagation has become the standard method for training artificial neural networks, it is required for the activation function to be analytic \cite{hirose:2006:complexnn}, i.e. the derivative exists at any point. In the complex domain, by the Liouville theorem, it is known that a bounded function, which is analytical at any point, is a constant function; reciprocally, complex non-constant functions have non-bounded images \cite{spiegel:2006:complexvariables}. Accordingly, some common real-valued functions, like the hyperbolic tangent and sigmoid, will diverge to infinity when they are extended to the complex domain \cite{hirose:2006:complexnn}; this makes them unsuitable for representing the behavior of a biological neuron. A similar problem arises in the quaternion domain: ``the only quaternion function regular with bounded norm in $E^4$ is a constant" \cite{deavours:1973:qcalculus}, where $E^4$ stands for a 4-dimensional Euclidean space.

The relationship between regular functions and the existence of their quaternionic derivatives is stated by the Cauchy-Riemann-Fueter equation \cite{sudbery:1979:cauchyriemannfuetereq}; leading to the result that the only globally analytic quaternion functions are some linear and constant functions. Moreover, non-linear activation functions are required for constructing a neural network architecture that works as a universal interpolator of a continuous quaternion valued function \cite{arena:1997:qmlp2}.

In this manner, a typical approach to circumvent this problem, has been to relax the constraints, and use non-linear non-analytic quaternion functions satisfying input and output properties, where the learning dynamics is built using partial derivatives on the quaternion domain. An example of this approach are split quaternion functions, defined as follows: 
\begin{equation}
  f(\mathbf{q})=f_R(\mathbf{q})+f_I(\mathbf{q})\hat{i}+f_J(\mathbf{q})\hat{j}+f_K(\mathbf{q})\hat{k},
\end{equation}
where $\mathbf{q}\in\mathbb{H}$, $f_R$, $f_I$, $f_J$, and $f_K$  are mappings over the real numbers: $f_{*}:\mathbb{R}\rightarrow \mathbb{R}$.

Nowadays, split quaternion functions remain the only type of activation function that has been implemented on QCNNs. Even though any type of quaternion split activation function used in QMLP can be applied, split quaternion ReLU is currently the most common activation function applied on QCNNs. For the sake of completeness, we present the activation functions found in current works.

Let $\mathbf{q}=q_R+q_I\hat{i}+q_J\hat{j}+q_K\hat{k} \in \mathbb{H}$, we have the following activation functions:
\begin{enumerate}
  \item{Split Quaternion Sigmoid}. It was introduced in \cite{altamirano:2017:geometricperception, arena:1994:qmlp0}, and is defined as follows:
  \begin{equation}
    \mathbb{Q}S(\mathbf{q})= S(q_R)+ S(q_I) \hat{i} + S(q_J) \hat{j} + S(q_K) \hat{k},
  \end{equation}
  where $S:  \mathbb{R}\rightarrow \mathbb{R}$ is the real-valued sigmoid function:
  \begin{equation}
    S(x)=  \frac{1}{1+e^{-x}}.
  \end{equation}

  \item{Split Quaternion Hyperbolic Tangent}. It was introduced in \cite{parcollet:2019:qcnnrecurrent, zhu:2018:qcnn}, and is defined as follows:
  \begin{eqnarray}
    \mathbb{Q}tanh(\mathbf{q}) &=& \tanh(q_R) + \tanh(q_I)\hat{i} \nonumber\\
    && + \tanh(q_J) \hat{j} \nonumber\\
    && + \tanh(q_K) \hat{k},
  \end{eqnarray}
  where $\tanh:  \mathbb{R}\rightarrow \mathbb{R}$ is the real-valued hyperbolic tangent function:
  \begin{eqnarray}
    \tanh(x)    &=& \frac{\sinh(x)}{\cosh(x)} \nonumber \\
    &=& \frac{\exp(2x)-1}{\exp(2x)+1}.
  \end{eqnarray}

  \item {Split Quaternion Hard Hyperbolic Tangent}. It was introduced in \cite{parcollet:2019:qcnn}, and is defined as follows:
  \begin{eqnarray}
    \mathbb{Q}H^2T(\mathbf{q}) &=& H^2T(q_R) + H^2T(q_I) \hat{i} \nonumber \\
    && + H^2T(q_J) \hat{j} \nonumber \\
    && + H^2T(q_K) \hat{k},
  \end{eqnarray}
  where $H^2T: \mathbb{R}\rightarrow \mathbb{R}$ \cite{collobert:2004:hardtanh}:
  \begin{equation}
    H^2T(x)= 
    \begin{cases}
      -1 & \text{si } x<-1 \\
      x & \text{si } -1\leq x\leq 1 \\
      1 & \text{si } x>1.
    \end{cases}
  \end{equation}
  For this type of activation functions, Collobert \cite{collobert:2004:hardtanh} proved that hidden layers of MLP works as local SVM's when the real-value hard hyperbolic tangent is used.

  \item{Split Quaternion ReLU}. It was introduced in \cite{gaudet:2018:qcnn, hongo:2020:qcnn, yin:2019:qcnn}, and is defined as follows:
  \begin{eqnarray}
    \mathbb{Q}ReLU(\mathbf{q}) &=& ReLU(q_R) + ReLU(q_I)\hat{i} \nonumber \\
    && + ReLU(q_J)\hat{j} \nonumber \\
    && + ReLU(q_K)\hat{k},
  \end{eqnarray}
  where $ReLU:  \mathbb{R}\rightarrow \mathbb{R}$ is the real-valued ReLU function \cite{fukushima:1969:relu, goodfellow:2016:deeplearning}:
  \begin{equation}
    ReLU(x)= max(0, x).
  \end{equation}

  \item{Split Quaternion Parametric ReLU}. It was introduced in \cite{parcollet:2018:qcnn}, and is defined as follows:
  \begin{eqnarray}
    \mathbb{Q}PReLU(\mathbf{q}) &=& PReLU(q_R)+ PReLU(q_I)\hat{i} \nonumber\\
    && +PReLU(q_J)\hat{j} \nonumber\\
    && + PReLU(q_K)\hat{k},
  \end{eqnarray}
  where $PReLU: \mathbb{R}\rightarrow \mathbb{R}$ \cite{he:2015:weightinicialization}:
  \begin{equation}
    PReLU(x)= 
    \begin{cases}
      x & \text{si } x>0 \\
      \alpha x & \text{si } x\leq 0
    \end{cases}
  \end{equation}
  and $\alpha$ is a parameter learned during the training stage, which controls the slope of the negative side of the ReLU function. Equivalently, we have:
  \begin{equation}
    PReLU(x)= max(0, x)+\alpha min(0,x).
  \end{equation}

  \item{Split Quaternion Leaky ReLU}. It was introduced in \cite{grassucci:2021:qcnn} as a particular case of the $\mathbb{Q}$PReLU function; in this case, $\alpha$ is a small constant value, e.g. $0.01$ \cite{maas:2013:leakyrelu}.
\end{enumerate}

An advantage of using split activation functions is that by processing each channel separately, we can adopt existing frameworks without additional modifications of the source code; however, this separate processing does not adequately capture the cross-dynamics of the data channels.

A different approach in the design of quaternion activation functions comes from quaternion analysis \cite{sudbery:1979:cauchyriemannfuetereq}. So far, it is clear that the key problem is designing suitable activation functions, and computing its quaternion derivatives; thus, some mathematicians have been working in redefining quaternion calculus. 

For example, De Leo and Rotelli \cite{deleo:1997:quaternioncalculus, deleo:2003:quaternioncalculus}, as well as Schwartz \cite{schwartz:2009:quaternioncalculus}, have introduced the concept of local quaternionic derivative. The trick was to  extend ``the concept of a derivative operator from those with constant (even quaternionic) coefficients, to one with variable coefficients depending upon the point of application... [thus] the derivative operator passes from a global form to local form" \cite{deleo:2003:quaternioncalculus}. Using this novel approach, they define \textit{local analyticity}, which, in contrast to global analyticity, does not reduce functions to a trivial class (constant or linear functions). Following this idea, Ujang \textit{et al.} \cite{ujang:2011:qtanh} define the concept of fully quaternion functions, and the properties they should fulfill to be locally analytic, non-linear, and suitable for gradient-based learning. They displayed better performance over split-quaternion functions when they were applied for designing adaptive filters. In the same train of thought, T. Isokawa proposed a QMLP and a backpropagation algorithm \cite{isokawa:2012:qmlp, isokawa:2013:qmlp}. Opposed to split functions, fully quaternion functions capture interchannel relationships, making them suitable for quaternion-based learning; however, experimental comparison over a standard benchmark remain an open issue.

To the best of our knowledge, currently, the only fully quaternion activation function that has been applied in QCNNs is the \textit{Rotation-Equivariant ReLU} function. It was proposed by Shen \textit{et al.} \cite{shen:2020:qcnn} as part of a model that extracts rotation equivariant features. Let $\{\mathbf{q_1}, \mathbf{q_2}, \dots, \mathbf{q_N}\}$ be a set of quaternions; then, for a quaternion $\mathbf{q_k}$, with $1\leq k\leq N$, the activation function is defined as follows:
\begin{equation}
  \mathbb{Q}REReLU (\mathbf{q_k})= \frac{\|\mathbf{q_k}\|}{max(\|\mathbf{q_k}\|,c)}\mathbf{q_k};
\end{equation}
where $c$ is a positive constant, computed as follows: 
\begin{equation}
  c=\frac{1}{N}\sum_{j=1}^N \|\mathbf{q_j}\|.
\end{equation}

In the same trend of developing novel quaternion calculus tools, but from a different perspective, Mandic \textit{et al.} \cite{mandic:2011:quaternionfunctions} start from the observation that for gradient-based optimization, a common objective is to minimize a positive real function of quaternion variables, e.g. $J(\mathbf{e},\mathbf{\bar{e}})=\mathbf{e}\mathbf{\bar{e}}$, and for that purpose it is used the pseudo-gradient, i.e. the sum of component-wise gradients. Formalization of these ideas is achieved by ``establishing the duality between derivatives of quaternion valued functions in $\mathbb{H}$ and the corresponding quadrivariate real functions in $\mathbb{R}^4$" \cite{mandic:2011:quaternionfunctions}, leading to what is called Hamiltonian-Real Calculus (HR). In addition, Mandic \textit{et al.} \cite{mandic:2011:quaternionfunctions} proved that for a real function of quaternion vector variable, the maximum change is in the direction of the conjugate gradient, establishing a general framework for quaternion gradient-based optimization. Going further, Xu and Mandic \cite{xu:2015:ghrcalculus} proposed the product and chain rules for computing derivatives, as well as the quaternion counterparts of the mean value and Taylor's theorems; this establishes an alternative framework called Generalize HR calculus (GHR). Thereafter, novel quaternion gradient algorithms using GHR calculus were proposed \cite{xu:2015:ghrapps, xu:2016:ghrapps}. Although in \cite{xu:2017:quaternionfunctions} is shown that a QMLP trained with a GHR-based algorithm obtains better prediction gains, on the 4D Saito's chaotic signal task, than other quaternion-based learning algorithms \cite{arena:1997:qmlp2, buchholz:2008:qnn, matsui:2004:geometricquaternionnn}, further experimental analysis on a standard benchmark and proper comparison with real and complex counterparts is required. To this date, there is no published work on the use of QRH calculus on QCNNs.

\textit{Future directions:} So far, we have identified the fundamental trends of thought for activation functions: split-quaternion functions, whose derivatives for training are computed in a channel wise manner, and fully quaternion functions, whose derivatives are computed locally, or using partial derivatives. Theoretical analysis, as well as some preliminary experimental results, have indicated a better performance of fully quaternion activation functions over others, as well as better performance of quaternion training methods based on GHR calculus. However, these ideas have not been set in practice on QCNNs. Intended works should focus on introducing novel fully connected activation functions that exploit specific properties of the quaternion representation, and establishing proper benchmarks for comparison of existing functions and training methods. In respect to quaternion calculus, the following section will connect the preceding ideas with the training of QCNNs and its future work.

\subsection{Training}

Currently, all of the methods for training that have been tested on QCNNs, rely on adaptations of the QMLP backpropagation algorithm \cite{arena:1996:qmlp1}, or an extension of the generalized complex chain rule for real-valued loss functions \cite{trabelsi:2018:complexnn}. Both approaches are equivalent and relax the analyticity condition by using partial derivatives with respect to the real and imaginary parts. Alternative approaches, such as GHR calculus or local derivatives have not been tested in current implementations of QCNNs.

Thus, Gaudet and Maida \cite{gaudet:2018:qcnn} introduced the Generalized Quaternion Chain Rule for a Real-valued function: Let $L$ be a real-valued loss function and $\mathbf{q}=q_R+q_I\hat{i}+ q_J\hat{j} + q_K\hat{k}$, where $q_R, q_I, q_J, q_K \in \mathbb{R}$, then \cite{gaudet:2018:qcnn, sudbery:1979:cauchyriemannfuetereq}: 
\begin{equation}
  \frac{\partial L}{\partial \mathbf{q}} = \frac{\partial L}{\partial q_R}+ \frac{\partial L}{\partial q_I}\hat{i} + \frac{\partial L}{\partial q_J}\hat{j} + \frac{\partial L}{\partial q_K}\hat{k}.
\end{equation}
Now, assuming $\mathbf{q}$ can be expressed in terms of a second quaternion variable, $\mathbf{g}=g_R+g_I \hat{i}+ g_J \hat{j} + g_K \hat{k}$, where $g_R, g_I, g_J, g_K \in \mathbb{R}$. Then, the chain rule is calculated as follows \cite{gaudet:2018:qcnn}: 
\begin{eqnarray}
  \frac{\partial L}{\partial \mathbf{g}} &=& \frac{\partial L}{\partial q_R} \left( \frac{\partial q_R}{\partial g_R} +\frac{\partial q_R}{\partial g_I}\hat{i} + \frac{\partial q_R}{\partial g_J}\hat{j} + \frac{\partial q_R}{\partial g_K}\hat{k} \right) + \nonumber \\
  && \frac{\partial L}{\partial q_I} \left( \frac{\partial q_I}{\partial g_R} +\frac{\partial q_I}{\partial g_I}\hat{i} + \frac{\partial q_I}{\partial g_J}\hat{j} + \frac{\partial q_I}{\partial g_K}\hat{k} \right) + \nonumber \\
  && \frac{\partial L}{\partial q_J} \left( \frac{\partial q_J}{\partial g_R} +\frac{\partial q_J}{\partial g_I}\hat{i} + \frac{\partial q_J}{\partial g_J}\hat{j} + \frac{\partial q_J}{\partial g_K}\hat{k} \right) + \nonumber \\
  && \frac{\partial L}{\partial q_K} \left( \frac{\partial q_K}{\partial g_R} +\frac{\partial q_K}{\partial g_I}\hat{i} + \frac{\partial q_K}{\partial g_J}\hat{j} + \frac{\partial q_K}{\partial g_K}\hat{k} \right)
\end{eqnarray}

These equations are applied in the implementation of the backpropagation algorithm. Since this is the most used approach, we call it the \textit{Standard Quaternion Backpropagation Algorithm} \cite{altamirano:2017:geometricperception, arena:1994:qmlp0, arena:1998:qbackprop}, summarized as follows:

 Let $\mathbf{x}\in\mathbb{H}$ be the input to a layer $C$, $\mathbf{w_{nm}}\in\mathbb{H}$ represents the weight connecting input $n$ to output $m$, and $\mathbf{d}^{top}, \mathbf{d}^{bottom} \in\mathbb{H}$ are the error propagated from the top and to the bottom layers, respective, and $\epsilon\in\mathbb{R}$ be the learning rate; then, for the current convolution layer, $C$:
\begin{enumerate}
  \item Update its weights using the following equation:
  \begin{equation} \label{eq:backprop_w}
    \mathbf{w_{nm}}=\mathbf{w_{nm}} + \epsilon \mathbf{d_m^{top}} \mathbf{\bar{x}_n}
  \end{equation}
  \item Update the bias term:
  \begin{equation} \label{eq:backprop_b}
    \mathbf{b_m}=\mathbf{b_m} + \epsilon  \mathbf{d_m^{top}}
  \end{equation}
  \item Propagate the error to the bottom layer according to the following equation:
  \begin{equation} \label{eq:backprop_delta}
    \mathbf{d_n^{bottom}} =  \sum_m (\mathbf{\bar{w}_{nm}} \mathbf{d_m^{top}})
  \end{equation}
\end{enumerate}

Note that Equations (\ref{eq:backprop_w}) and (\ref{eq:backprop_delta}) use quaternion products. For an activation layer, the error is propagated to the bottom layer according to the following equation:
\begin{equation}  \label{eq:backprop_actfn}
  \mathbf{d_n}^{bottom} =   \mathbf{d_n}^{top} \odot f'(\mathbf{x_n})
\end{equation}
where $\odot$ is the Hadamard or Schur product (component wise).

Adopting the same definition of the chain rule, but a different definition of convolution, we have the work of Zhu \textit{et al.} \cite{zhu:2018:qcnn}, whom apply the two-sided convolution and a polar representation of the quaternion weights. Since their model applies rotation and scaling on the input features, the quaternion gradient for his model is simplified to a rotation transformation over the same axis, but with a reversed angle. A general model of this approach, with learnable arbitrary axes, is presented by \cite{matsumoto:2022:fullyrotationalqcnn}. Therefore, the backpropagation algorithm is similar to the one presented before, but Equation (\ref{eq:backprop_w}) changes accordingly \cite{isokawa:2009:quaternionnn, matsui:2004:geometricquaternionnn, matsumoto:2022:fullyrotationalqcnn}:
\begin{eqnarray}
  \mathbf{w_{nm}} = \mathbf{w_{nm}} + \nonumber\\
  \frac{\epsilon}{\|\mathbf{w_{nm}}\|}
  \left\{ \frac{\mathbf{d_m^{top}} \cdot \left( \mathbf{w_{nm}} \mathbf{x_n} \mathbf{\bar{w}_{nm}} \right)} {\|\mathbf{w_{nm}}\|^2} \mathbf{w_{nm}} - 2\mathbf{d_m^{top}} \mathbf{w_{nm}} \mathbf{\bar{x}_n} \right\} 
\end{eqnarray}
and Equation (\ref{eq:backprop_delta}) changes to \cite{isokawa:2009:quaternionnn, matsui:2004:geometricquaternionnn, matsumoto:2022:fullyrotationalqcnn}:
\begin{equation}
  \mathbf{d_n^{bottom}} =  \sum_m \frac{ \mathbf{\bar{w}_{nm}}  \mathbf{d_m^{top}} \mathbf{w_{nm}}}{\| \mathbf{w_{nm}} \|}
\end{equation}

From the point of view of supervised learning, the problem of training a QCNN can be stated as an optimization one:
\begin{equation}
  \underset{\mathbf{w_1},\dots,\mathbf{w_n}}{\mathrm{argmin}}\ L(\mathbf{w_1},\dots,\mathbf{w_n})
\end{equation}
where $L$ is the loss function, and $\mathbf{w_1},\dots,\mathbf{w_n}$ are all the quaternion weights of the network.

Besides the problem of computing derivatives of non-analytic quaternion activation functions; another problem is that algorithms such as gradient descendant and back propagation could be trapped in local minima. Therefore, QCNNs could be trained with alternative methods that not rely on the computing of quaternion derivatives, such as evolutionary algorithms, ant colony optimization, particle swam optimization, etc. \cite{chong:2021:metaheuristicstraining}. In this case, general-purpose optimization algorithms will have the same performance in average \cite{1995:wolpert:nofreelunch, wolpert:1997:nofreelunch}.

\textit{Future directions:} Novel methods for training should be applied, for example: modern quaternion calculus techniques, such as local or GHR calculus, and meta-heuristic algorithms working in the quaternion domain. In addition, fully quaternion loss functions should be introduced.

\subsection{Quaternion weight initialization}

At the beginning of this century, real-valued neural networks were showing the superiority of deep architectures; however, the standard gradient descent algorithm from random weight initialization performed poorly when used for training these models. To understand the reason of this behavior, Glorot and Bengio \cite{glorot:2010:weightinicialization} established an experimental setup and observed that some activation functions can cause saturation in top hidden layers. In addition, by theoretically analyzing the forward and backward propagation variances (expressed with respect to the input, output and weight initialization randomness), they realized that, because of the multiplicative effect through layers, the variance of the back-propagated gradient might vanish or explode in very deep networks. Consequently, they proposed, and validated experimentally, a weight initialization procedure, which makes the variance dependent on each layer, and maintains activation and back-propagated gradient variances as we move up or down the network. Their method, called \textit{normalized initialization}, uses a scaled uniform distribution for initialization. However, one of its assumptions is that the network is in a linear regime at the initialization step; thus, this method works better for softsign units than for sigmoid or hyperbolic tangent ones \cite{glorot:2010:weightinicialization}. Since the linearity assumption is invalid for activation functions such as ReLU, He \textit{et al.} \cite{he:2015:weightinicialization} developed an initialization method for non-linear ReLU and Parametric ReLU activation functions. Their initialization method uses a zero-men Gaussian distribution with a specific standard deviation value, and surpasses the performance of the previous method for training extremely deep models, e.g. 30 convolution layers.

In the case of classic QCNNs, Trabelsi \textit{et al.} \cite{trabelsi:2018:complexnn} extended these results to deep complex networks. Similar results are presented by Gaudet and Maida \cite{gaudet:2018:qcnn} for deep quaternion networks, and by Parcollet \textit{et al.} \cite{parcollet:2019:qcnnrecurrent} for quaternion recurrent networks. These works treat quaternion weights as 4-dimensional vectors, whose components are normally distributed, centered at zero, and independent. Hence, it can be proved that the weights and their magnitude follow a 4DOF Rayleigh distribution \cite{gaudet:2018:qcnn, parcollet:2019:qcnnrecurrent}, reducing the weight initialization by selecting a single parameter, $\sigma$, which indicate the mode of the distribution. If $\sigma=\frac{1}{\sqrt{2(n_{in}+n_{out})}}$ we have a \textit{quaternion normalized initialization} which ensures that the variances of the quaternion input, output and their gradients are the same; while $\sigma=\frac{1}{\sqrt{2n_{in}}}$ is used for the \textit{quaternion ReLU initialization}, where  $n_{in}$, and $n_{out}$ are the number of neurons of the input and output layers, respectively. The method is presented in Algorithmic \ref{alg:weightinitialization}.

\begin{algorithm}[!t]
  \caption {Quaternion-valued weight initialization} \label{alg:weightinitialization}
  \begin{algorithmic}[1]
    \REQUIRE $\mathbf{W}\in\mathbb{H}^{n_{in}\cdot n_{out}}, n_{in}\in\mathbb{N}^+, n_{out}\in\mathbb{N}^+$
    \IF {RELU} \STATE {$\sigma\gets\frac{1}{\sqrt{2n_{in}}}$} \ELSE \STATE{$\sigma\gets\frac{1}{\sqrt{2(n_{in}+n_{out})}}$} \ENDIF
    \FORALL{$\mathbf{w}$ in $\mathbf{W}$}
      \STATE {$\phi\gets rayleight\_rand(\sigma)$}
      \STATE {$\theta\gets uniform\_rand(-\pi,\pi)$} 
      \STATE {$x,y,z\gets uniform\_rand(0,1)$}
      \STATE {$\hat{u}\gets \frac{x\hat{i}+y\hat{j}+z\hat{k}}{\sqrt{x^2+y^2+z^2}}$}
      \STATE {$\mathbf{w}\gets \phi\cos(\theta)+\phi\sin(\theta)\hat{u}$}
    \ENDFOR\\
    \RETURN $\mathbf{W}$
  \end{algorithmic}
\end{algorithm}

On the other hand, for geometric QCNNs, the quaternion weights represent an affine transformation composed by a scale factor, $s$, and a rotation angle $\theta$. Thus, to keep the same variance of the gradients during training, Zhu \textit{et al.} \cite{zhu:2018:qcnn} proposed a simple initialization procedure using the uniform distribution, $U[\cdot]$:
\begin{eqnarray}
  s_j &\sim& U \left[-\frac{\sqrt{6}}{\sqrt{n_j+n_{j+1}}}, \frac{\sqrt{6}}{\sqrt{n_j+n_{j+1}}} \right] \nonumber\\
  \theta &\sim& U\left[-\frac{\pi}{2}, \frac{\pi}{2} \right]
\end{eqnarray}

\textit{Future directions:} In the case of geometric QCNNs, further theoretical analysis of weight initialization techniques is required, as well as the extension for ReLU units.
For equivariant QCNNs, weight initialization procedures have not been introduced or or are not described in the literature. In addition, current works only consider the case of split quaternion activation functions. The propagation of variance using fully quaternion activation functions should be investigated.

\section{Architectural design for applications.}\label{sec:apps}

The previous section deals with the individual building blocks of QCNNs; the possible ways we can interconnect them, give rise to numerous models. In this manner, there are three leading factors to consider when implementing applications:
\begin{enumerate}
  \item Domain of application. Current works are primarily focused on 3 areas: vision, language, and forecasting.
  \item Mapping the input data from real numbers to the quaternion domain. 
  \item Topology. Based on current works, we have the following types:
  \begin{itemize}
    \item ConvNets. They use convolution layers without additional tricks.
    \item Residual. They use a shortcut connection from input to forward blocks, allowing to learn with reference to the layer inputs instead of learning unreferenced functions.
    \item Convolution Auto-Encoders (CAE). Its aim is to reconstruct the input feature at the output.
    \item Point-based. They focused on unordered sets of vectors such as point-cloud input data.
    \item Recurrent. They exploit connections that can create cycles, allowing the network to learn long-term temporal information.
    \item Generative. They use an adversarial learning paradigm, where one player generates samples resembling the real data, and the other discriminates between real and fake data; the solution of the game is the model achieving Nash equilibrium.
  \end{itemize}
\end{enumerate}

This section is devoted to show how the blocks presented in Section \ref{sec:qcnnblocks} can be used to construct different architectures, and we focus our analysis on the three points previously introduced.

The information is organized in three subsections corresponding to the domains of applications: vision, language, and forecasting. Within each subsection, the models are ordered according to the type of convolution they applied: classic, geometric, or equivariant. In addition, different networks are presented according to their topology: ConvNet, Residual, CAE, Point-based, Recurrent, or Generative. Methods for mapping input data to the quaternion domain are presented in each subsection.

\begin{figure*}[!t]
  \centering
  \includegraphics[width=0.66\textwidth]{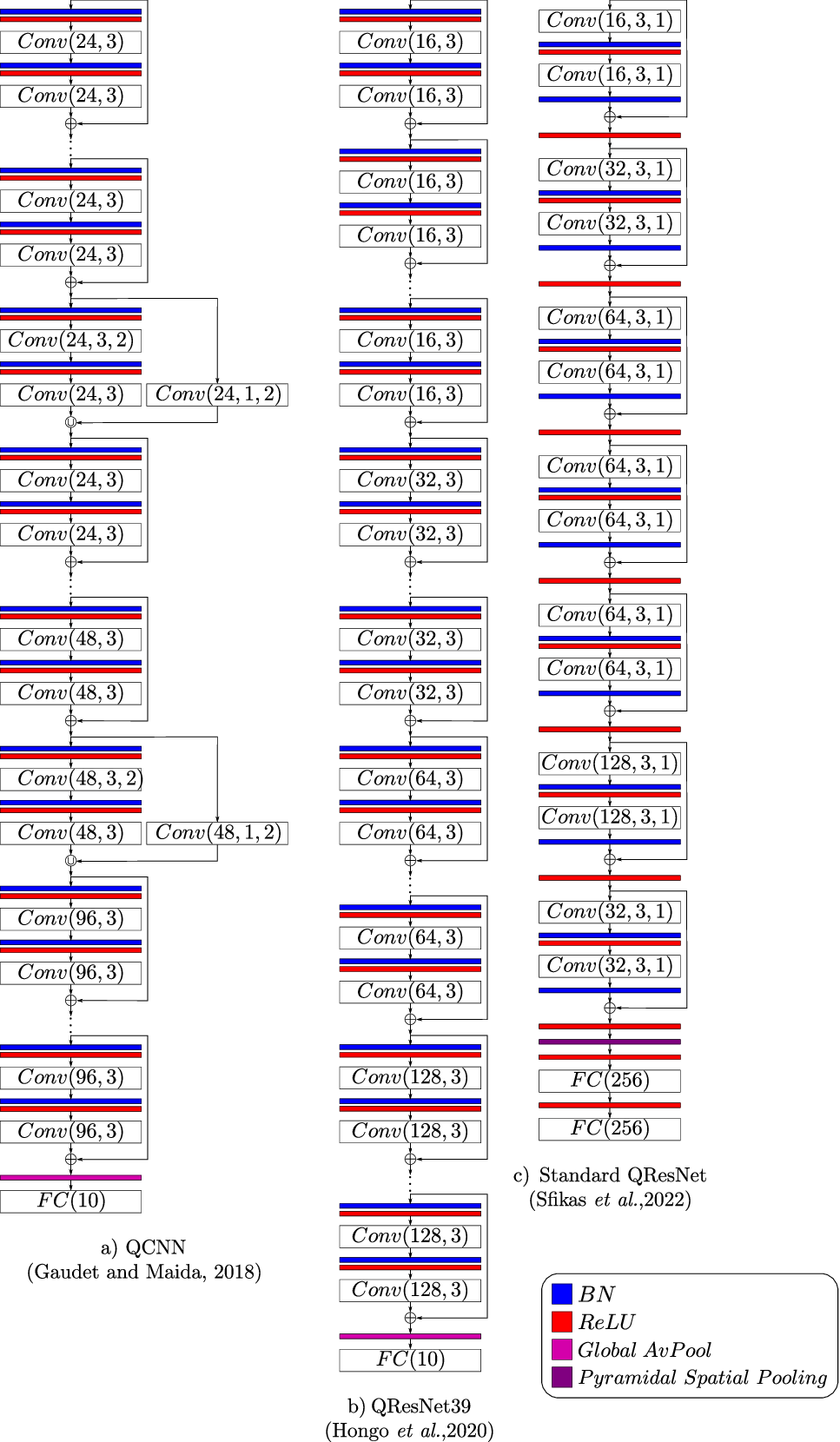}
  \caption {Residual QNN models. Some notation details: For convolution, parameters are presented in the format (number of kernels, size of kernel, stride); if the size of kernel or stride is the same for each dimension, it is shown just a single number, e.g. instead of $3\times 3$ it is shown $3$. Symbol $+$ inside a circle means summing. Dropout and flatten procedures were omitted. All blocks represent operations in the quaternion domain.}
  \label{fig:resqcnn}
\end{figure*}

\begin{figure*}[!t]
  \centering
  \includegraphics[width=0.70\textwidth]{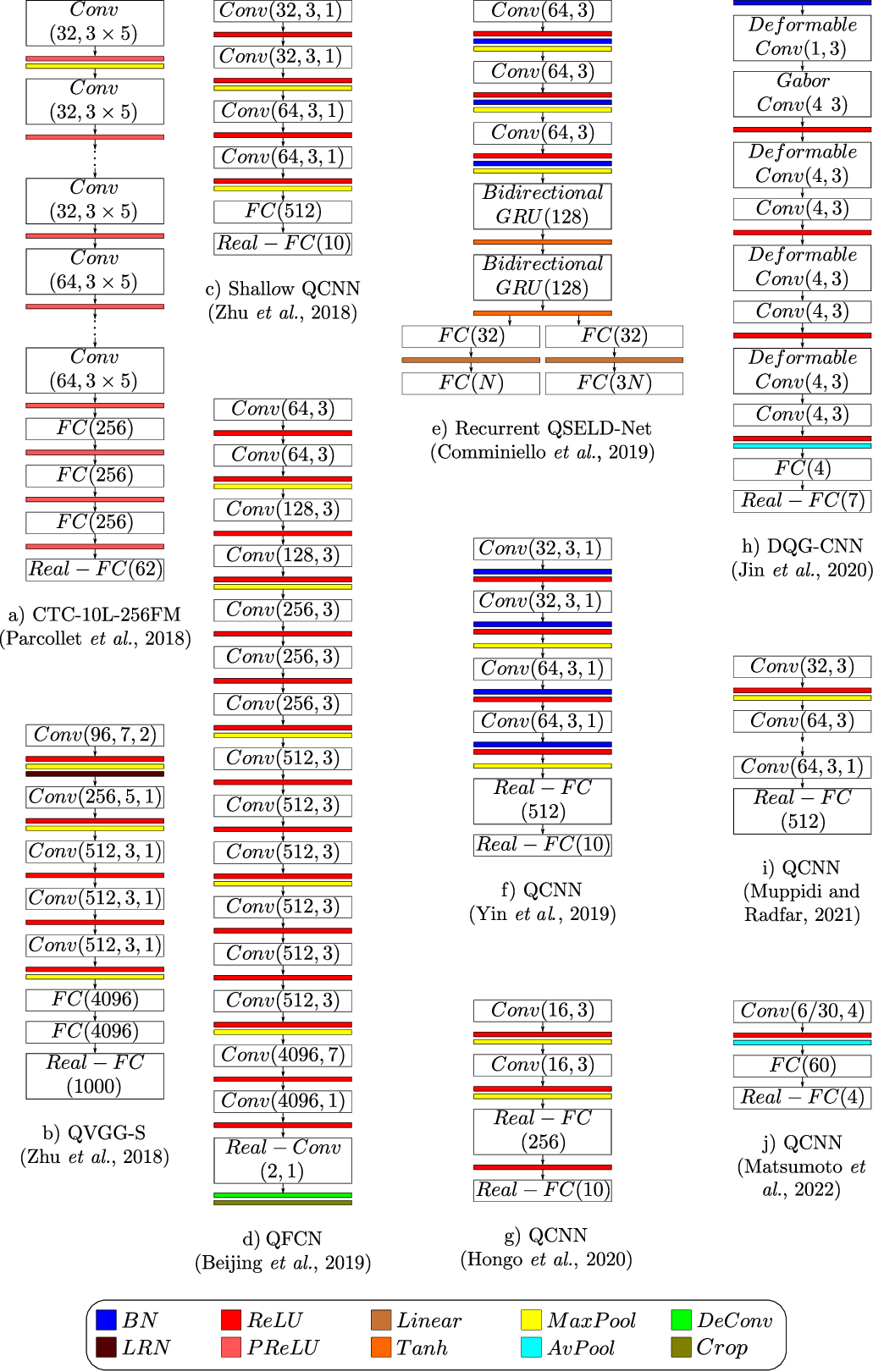}
  \caption {Quaternion ConvNet models. Some notation details: For convolution, parameters are presented in the format (number of kernels, size of kernel, stride); if the size of kernel or stride is the same for each dimension, it is shown just a single number, e.g. instead of $3\times 3$ it is shown $3$. LRN stands for Local Response Normalization. Dropout and flatten procedures were omitted. All blocks represent operations in the quaternion domain, except those preceded by the word ``real".}
  \label{fig:qconvnets}
\end{figure*}

\begin{figure*}[!t]
  \centering
  \includegraphics[width=0.70\textwidth]{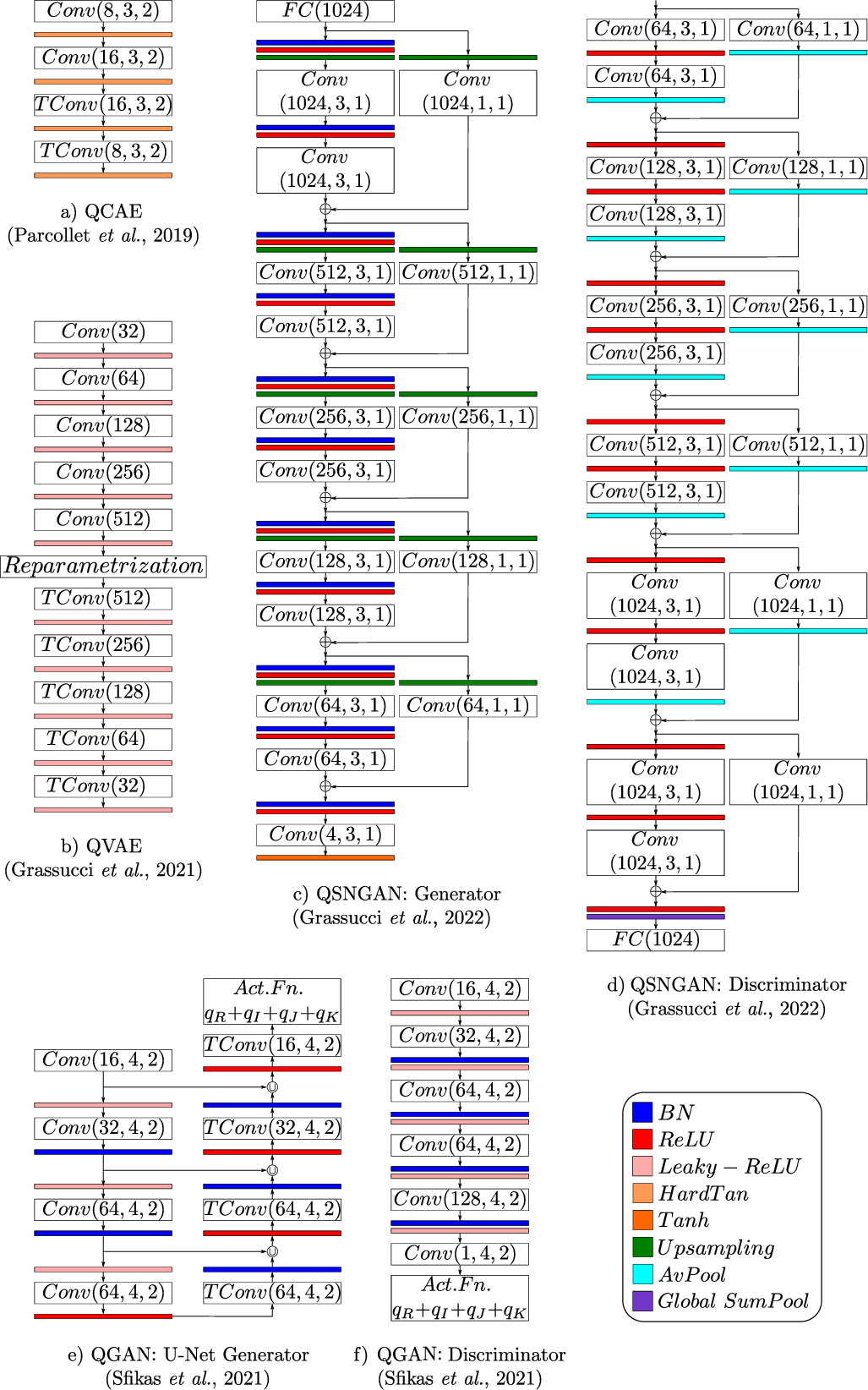}
  \caption {Quaternion CAE and Generative models. Some notation details: For convolution, parameters are presented in the format (number of kernels, size of kernel, stride); if the size of kernel or stride is the same for each dimension, it is shown just a single number, e.g. instead of $3\times 3$ it is shown $3$. Symbol $+$ inside a circle means summing while $\cup$ inside a circle means concatenation. TConv stands for Decoding Transposed Convolution. Dropout and flatten procedures were omitted. All blocks represent operations in the quaternion domain, except those preceded by the word ``real".}
  \label{fig:qgen}
\end{figure*}

A graphic depiction of several models is shown in Figures \ref{fig:resqcnn}, \ref{fig:qconvnets}, and \ref{fig:qgen}. Some architectures available in the literature are presented from a high-level abstraction point of view; showing their topologies and the blocks they use. Information about the number of quaternion kernels, size, and stride was included when available. Further implementation details, such as weight initialization, optimization methods, etc. were omitted to avoid a cumbersome presentation. Readers interested in specific models are encouraged to consult the original sources.

In addition, Tables~\ref{tab:appqcnns1} to \ref{tab:appqcnns4} provide a comparison between different QCNNs. They provide extra information, such as datasets on which the models were tested, performance metrics, comparison to real-valued models, etc. For comparison between quaternion and real-valued models, we only report real-valued architectures that have similar topology. The difference in the number of parameters between real-valued and quaternion-valued networks is due to the following reasons: Some authors prefer to compare networks with the same number of real-valued units, and quaternion-valued units, where the quaternion-valued units have $4$ times more parameters. Others compare networks with the same number of parameters: they reduce the number of quaternion units to $25\%$ of the real-valued networks, or quadruple the number of real-valued units. In addition, some quaternion-value networks use real-valued components in some parts of the network. 

To conclude, Subsection \ref{subsec:insights} summarizes some insights obtained from this works.

\begin{table*}[h!]
  \caption{Applications of QCNNs in Computer Vision, Part I. The meaning of acronyms used in the performance column is the following: IoU (Intersection over Union $\uparrow$), SSIM (Structural Similarity $\uparrow$), MSE (Mean Square Error $\downarrow$), FID (Fréchet Inception Distance $\downarrow$), BCE (Binary Cross Entropy $\downarrow$), MAP (Mean Average Precision $\uparrow$), CA (Classification Accuracy $\uparrow$). The $\uparrow$ means the higher the better, while the $\downarrow$ means the lower the better.} \label{tab:appqcnns1}
  \begin{tabular*}{
    \textwidth}{@{\extracolsep{\fill}} p{0.133\textwidth} p{0.11\textwidth} p{0.125\textwidth} p{0.095\textwidth} p{0.04\textwidth} p{0.091\textwidth} p{0.04\textwidth} p{0.091\textwidth} @{\extracolsep{\fill}}}
    \hline
     {Authors} & {Arch type} & {Problem} & {Dataset} & \multicolumn{2}{c}{Quaternion model} & \multicolumn{2}{c}{ Real-valued model} \\
    \cline{5-8}
    {(Conceptual trend)} & {(Model)} & & & {No. Param.} & {Perform.} & {No. Param.} & {Perform.} \\
    \hline
    Gaudet and Maida \cite{gaudet:2018:qcnn} (Classic) & Residual (Deep QCNN) & Image classification & CIFAR-10 \cite{krizhevsky:2009:cifar} & 932.8K & CA: 94.56\% & 3.6M & CA: 93.63\% \\

    Gaudet and Maida \cite{gaudet:2018:qcnn} (Classic) & Residual (Shallow QCNN) & Image classification & CIFAR-10 \cite{krizhevsky:2009:cifar} & 133.6K & CA: 93.23\% & 508.9K & CA: 93.18\% \\

    Hongo \textit{et al.} \cite{hongo:2020:qcnn} (Geometric) & Residual (ResNet39) & Image classification & CIFAR-10 \cite{krizhevsky:2009:cifar} & N.A. & CA: 93-94\% & N.A. & CA: 93-94\% \\

    Yin \textit{et al.} \cite{yin:2019:qcnn} (Classic) & ConvNet (Attentional QCNN) & Image classification & CIFAR-10 \cite{krizhevsky:2009:cifar} & N.A. & CA: 85.37\% &  N.A. & CA: 75.46\% \\

    Yin \textit{et al.} \cite{yin:2019:qcnn} (Classic) & ConvNet (QCNN) & Image classification & CIFAR-10 \cite{krizhevsky:2009:cifar} & N.A. & CA: 84.15\% &  N.A. & CA: 75.46\% \\

    Hongo \textit{et al.} \cite{hongo:2020:qcnn} (Geometric) & ConvNet (QCNN) & Image classification & CIFAR-10 \cite{krizhevsky:2009:cifar} & N.A. & CA: 80-82\% & N.A. & CA: 79-80\% \\

    Zhu \textit{et al.} \cite{zhu:2018:qcnn} (Geometric) & ConvNet (Shallow QCNN) & Image classification & CIFAR-10 \cite{krizhevsky:2009:cifar} & N.A. & CA: 77.78\% & N.A. & CA: 75.46\% \\

    Gaudet and Maida \cite{gaudet:2018:qcnn} (Classic) & Residual (Deep QCNN) & Image classification & CIFAR-100 \cite{krizhevsky:2009:cifar}  & 932.8K & CA: 73.99\% & 3.6M & CA: 71.93\% \\

    Gaudet and Maida \cite{gaudet:2018:qcnn} (Classic) & Residual (Shallow QCNN) & Image classification & CIFAR-100 \cite{krizhevsky:2009:cifar} & 133.6K & CA: 69.41\% & 508.9K & CA: 67.98\% \\

    Gaudet and Maida \cite{gaudet:2018:qcnn} (Classic) & Residual (Shallow QCNN) & Image classification & KITTI Road Est. \cite{fritsch:2013:kittiroadestimationbenchmark} & 128.7K &  IoU: 0.827 & 507K &  IoU: 0.827 \\

    Zhu \textit{et al.} \cite{zhu:2018:qcnn} (Geometric) & ConvNet (VGG-S) & Image classification & 102 Oxford flower \cite{nilsback:2008:oxfordflowers} & N.A. & CA: 76.95\% & N.A. & CA: 73.08\% \\

    Sfikas \textit{et al.} \cite{sfikas:2021:qgan} (Classic) & Generative (QGAN) & Scene text detection & Byzantine monuments \cite{rhoby:2017:bizantinedataset, kordatos:2013:bizantinedataset} & 1.5M & BCE: 6.54; IoU: 45.4\% & 6.1M & BCE: 7.4; IoU: 51.9\% \\

    Sfikas \textit{et al.} \cite{sfikas:2021:qgan} (Classic) & Generative (QGAN) & Scene text detection & Byzantine monuments \cite{rhoby:2017:bizantinedataset, kordatos:2013:bizantinedataset} &  6.1M & BCE: 6.91; IoU: 44.9\% & 24.2M & BCE: 6.45; IoU: 52 \% \\

    Sfikas \textit{et al.} \cite{sfikas:2022:greekmanuscript} (Classic) & Residual (Std QResNet) & Keyword spotting & PIOP-DAS \cite{sfikas:2022:piopdasdataset} & 5.9M & MAP: 94.3\% & 23.7M & MAP: 94.6 \% \\

    Sfikas \textit{et al.} \cite{sfikas:2022:greekmanuscript} (Classic) & Residual (Std QResNet) & Keyword spotting & GRPOLY-DB \cite{gatos:2015:grpolydataset}  &  5.9M & MAP: 98.6 \% & 23.7M & MAP: 98.6 \% \\

    Matsumoto \textit{et al.} \cite{matsumoto:2022:fullyrotationalqcnn} (Geometric) & ConvNet (QCNN) & Pixel classification & Hokkaido-JAXA-ALOS & N.A. & Avg F-Score: 0.857 & N.A. & Avg F-Score: 0.838 \\

    Matsumoto \textit{et al.} \cite{matsumoto:2022:fullyrotationalqcnn} (Geometric) & ConvNet (QCNN) & Pixel classification & Hokkaido-JAXA-ALOS & N.A. & Avg F-Score: 0.919 & N.A. & Avg F-Score: 0.890 \\

    Jin \textit{et al.} \cite{jin:2020:deformablegaborqcnn} (Classic) & ConvNet (DQG-CNN) & Facial recognition & Oulu-CASIA \cite{taini:2008:oulucasiadataset} & N.A. & CA: $99.38\%$ & N.A. & CA: $96.88\%$ \\

    Jin \textit{et al.} \cite{jin:2020:deformablegaborqcnn} (Classic) & ConvNet (DQG-CNN) & Facial recognition & MMI \cite{pantic:2005:mmi:dataset} & N.A. & CA: $99.36\%$ & N.A. & CA: $95.24\%$ \\

    Jin \textit{et al.} \cite{jin:2020:deformablegaborqcnn} (Classic) & ConvNet (DQG-CNN) & Facial recognition & SFEW \cite{dhall:2011:sfewdataset} & N.A. & CA: $45.86\%$ & N.A. & CA: $38.9\%$ \\

    Grassucci \textit{et al.} \cite{grassucci:2021:qcnn} (Classic) & CAE (QVAE) & Face generation & CelebA \cite{liu:2015:celebfacesdataset} & 1.4M & SSIM: 0.8941; MSE: 0.0031; FID: 1775.7 & 3.8M & SSIM: 0.8492; MSE: 0.0047; FID: 195.7 \\
    \hline
  \end{tabular*}
\end{table*}

\begin{table*}[h!]
  \caption{Applications of QCNNs in Computer Vision, Part II. The meaning of acronyms used in the performance column is the following: CA (Classification Accuracy $\uparrow$). The $\uparrow$ means the higher the better, while the $\downarrow$ means the lower the better. Information about the number of parameters was not found for this models.}
 \label{tab:appqcnns2}
  \begin{tabular*}{
    \textwidth}{@{\extracolsep{\fill}} p{0.158\textwidth} p{0.16\textwidth} p{0.127\textwidth} p{0.115\textwidth} p{0.118\textwidth} p{0.118\textwidth} @{\extracolsep{\fill}}}
    \hline
     {Authors} & {Arch type} & {Problem} & {Dataset} & {Quaternion model} & {Real model} \\
    \cline{5-6}
    {(Conceptual trend)} & {(Model)} & & & {Metrics} & {Metrics} \\
    \hline
    Shen \textit{et al.} \cite{shen:2020:qcnn} (Equivariant) & Point-based (PointNet++) & 3D shape classification & ModelNet 40 \cite{wu:2015:3dshapenetsdataset} &  CA: $63.95\%$ &  CA: $26.23\%$ \\

    Shen \textit{et al.} \cite{shen:2020:qcnn} (Equivariant) & Point-based (DG-CNN) & 3D shape classification & ModelNet 40 \cite{wu:2015:3dshapenetsdataset} &  CA: $83.08\%$ &  CA: $31.34\%$ \\

    Shen \textit{et al.} \cite{shen:2020:qcnn} (Equivariant) & Point-based (PointConv) & 3D shape classification & ModelNet 40 \cite{wu:2015:3dshapenetsdataset} &  CA: $78.14\%$ &  CA: $23.72\%$  \\

    Shen \textit{et al.} \cite{shen:2020:qcnn} (Equivariant) & Point-based (PointNet++) & 3D shape classification & 3D MNIST \cite{kaggke:2016:3dmnist} &  CA: $68.99\%$ &  CA: $51.16\%$  \\

    Shen \textit{et al.} \cite{shen:2020:qcnn} (Equivariant) & Point-based (DG-CNN) & 3D shape classification & 3D MNIST \cite{kaggke:2016:3dmnist} &  CA: $82.09\%$ &  CA: $49.25\%$ \\

    Shen \textit{et al.} \cite{shen:2020:qcnn} (Equivariant) & Point-based (PointConv) & 3D shape classification & 3D MNIST \cite{kaggke:2016:3dmnist} &  CA: $78.59\%$ &  CA: $50.95\%$ \\

    Beijing \textit{et al.} \cite{beijing:2019:fullyqcnn} (Classic) & ConvNet (FCN+ CRF) & Splicing localization & CASIA V1 \cite{dong:2013:casiadataset} & F-Measure: 0.505 & F-Measure: 0.484 \\

    Beijing \textit{et al.} \cite{beijing:2021:regionqcnn} (Classic) & ConvNet (Two-stream R-CNN) & Splicing localization & CASIA V1 \cite{dong:2013:casiadataset} &  F-Measure: 0.6721 &   N.A. \\

    Beijing \textit{et al.} \cite{beijing:2019:fullyqcnn} (Classic) & ConvNet (FCN+ CRF) & Splicing localization & CASIA V2 \cite{dong:2013:casiadataset} & F-Measure: 0.545 & F-Measure: 0.523 \\

    Beijing \textit{et al.} \cite{beijing:2021:regionqcnn} (Classic) & ConvNet (Two-stream R-CNN) & Splicing localization & CASIA V2 \cite{dong:2013:casiadataset} & F-Measure: 0.7787 &  N.A. \\

    Beijing \textit{et al.} \cite{beijing:2019:fullyqcnn} (Classic) & ConvNet (FCN+ CRF) & Splicing localization & DVMM \cite{ng:2004:dvmmdataset} & F-Measure: 0.488 & F-Measure: 0.451 \\

    Beijing \textit{et al.} \cite{beijing:2021:regionqcnn} (Classic) & ConvNet (Two-stream R-CNN) & Splicing localization & DVMM \cite{ng:2004:dvmmdataset} & F-Measure: 0.7939 &  N.A. \\

    \hline
  \end{tabular*}
\end{table*}

\begin{table*}[h!]
  \caption{Applications of QCNNs in Image Processing. The meaning of acronyms used in the performance column is the following: SSIM (Structural Similarity $\uparrow$), PSNR: (Peak Signal to Noise Ratio $\uparrow$), FID (Fréchet Inception Distance $\downarrow$), IS (Inception Score $\uparrow$), LPIPS (Learned Perceptual Image Patch Similarity $\uparrow$). The $\uparrow$ means the higher the better, while the $\downarrow$ means the lower the better.}\label{tab:appqcnns3}
  \begin{tabular*}{
    \textwidth}{@{\extracolsep{\fill}} p{0.133\textwidth} p{0.11\textwidth} p{0.085\textwidth} p{0.095\textwidth} p{0.04\textwidth} p{0.111\textwidth} p{0.04\textwidth} p{0.111\textwidth} @{\extracolsep{\fill}}}
    \hline
     {Authors} & {Arch type} & {Problem} & {Dataset} & \multicolumn{2}{c}{Quaternion model} & \multicolumn{2}{c}{ Real model} \\
    \cline{5-8}
    {(Conceptual trend)} & {(Model)} & & & {No. Param.} & {Metrics} & {No. Param.} & {Metrics} \\
    \hline

    Grassucci \textit{et al.} \cite{grassucci:2022:qsngan} (Classic) & Generative (SNGAN) & Image generation & CelebA-HQ \cite{karras:2018:celebfaceshqdataset} & 9.6M (35Gb) & FID: $33.068$; IS: $2.026\pm 0.082$ & 32.2M (115Gb) & FID: $34.483$; IS: $2.032\pm 0.062$ \\

    Grassucci \textit{et al.} \cite{grassucci:2022:starganv2} (Classic) & Generative (StarGANv2) & Image to image translation & CelebA-HQ \cite{karras:2018:celebfaceshqdataset}& 22M (76Mb) & Reference FID: $23.09$; Reference LPIPS: $0.22$; Latent FID: $27.90$; Latent LPIPS: $0.12$ & 87M (307Mb) & Reference FID: $21.24$; Reference LPIPS: $0.24$; Latent FID: $17.16$; Latent LPIPS: $0.25$ \\

     Grassucci \textit{et al.} \cite{grassucci:2022:qsngan} (Classic) & Generative (SNGAN) & Image to image translation & 102 Oxford flowers \cite{nilsback:2008:oxfordflowers} & 9.6M (35Gb) & FID: $115.838$; IS: $3\pm 0.141$ & 32.2M (115Gb) & FID: $165.058$; IS: $2.977\pm 0.146$ \\

    Parcollet \textit{et al.} \cite{parcollet:2019:qcnn} (Classic) & CAE (QCAE) & Grayscale to RGB images& KODAK PhotoCD \cite{franzen:2013:kodakdataset} & 6.4K & SSIM: 0.96; PSNR: 31.68 dB & 25K & SSIM: 0.87; PSNR: 29.95 dB \\
    
    Zhu \textit{et al.} \cite{zhu:2018:qcnn} (Geometric) & CAE (U-Net) & RGB image denoising & 102 Oxford flower \cite{nilsback:2008:oxfordflowers}& N.A. & PSNR: 31.318 dB & N.A. & PSNR: 30.979 dB \\
  
    Zhu \textit{et al.} \cite{zhu:2018:qcnn} (Geometric) & CAE (U-Net) & RGB image denoising & COCO \cite{tsung:2014:mscocodataset} & N.A. & PSNR:  30.726 dB & N.A. & PSNR: 30.490 dB  \\

    Yin \textit{et al.} \cite{yin:2019:qcnn} (Classic) & ConvNet (QCNN) & Double JPEG compression detection & UCID  \cite{huang:2010:uciddataset} & N.A. & Detection acc: $99.17-77.83\%$ &  N.A. & N.A. \\
    \hline
  \end{tabular*}
\end{table*}

\begin{table*}[h!]
  \caption{Applications of QCNNs in Natural Language Processing and Forecasting. The meaning of acronyms used in the performance column is the following: PER (Phoneme Error rate $\downarrow$), MSE (Mean Square Error $\downarrow$). The $\uparrow$ means the higher the better, while the $\downarrow$ means the lower the better.} \label{tab:appqcnns4}
  \begin{tabular*}{
    \textwidth}{@{\extracolsep{\fill}} p{0.133\textwidth} p{0.11\textwidth} p{0.105\textwidth} p{0.085\textwidth} p{0.06\textwidth} p{0.096\textwidth} p{0.04\textwidth} p{0.096\textwidth} @{\extracolsep{\fill}}}
    \hline
     {Authors} & {Arch type} & {Problem} & {Dataset} & \multicolumn{2}{c}{Quaternion model} & \multicolumn{2}{c}{ Real model} \\
    \cline{5-8}
    {(Conceptual trend)} & {(Model)} & & & {No. Param.} & {Metrics} & {No. Param.} & {Metrics} \\
    \hline
    Parcollet \textit{et al.} \cite{parcollet:2018:qcnn} (Classic) & ConvNet (CTC-10L-256FM) & Speech phoneme recognition & TIMIT \cite{1993:nist:timitdataset} & 8.1M & Test PER: $19.64\%$ & 32.1M & Test PER: $21.23\%$ \\

    Comminiello \textit{et al.} \cite{comminiello:2019:qcnn} (Classic) & ConvNet (Recurrent QSELD-Net) & Sound Event Localization and Detection & ANSYN \cite{adavanne:2018:seldnet} & 760K & Ave-O1: 0.123; Ave-O2: 0.313; Ave-O3: 0.437; & 530K & Ave-O1: 0.227; Ave-O2: 0.4; Ave-O3: 0.503 \\

    Comminiello \textit{et al.} \cite{comminiello:2019:qcnn} (Classic) & ConvNet (Recurrent QSELD-Net) & Sound Event Localization and Detection & RESYN \cite{adavanne:2018:seldnet} & 760K & Ave-O1: 0.25; Ave-O2: 0.433; Ave-O3: 0.51; & 530K & Ave-O1: 0.6; Ave-O2: 0.48; Ave-O3: 0.54 \\

    Muppidi and Radfar\cite{muppidi:2021:qcnn} (Geometric) & ConvNet (QCNN) & Emotion Classification & RAVDESS \cite{livingstone:2018:ravnessdataset} & 42 Mb & Acc: $77.87\%$ & N.A. & N.A. \\

    Muppidi and Radfar\cite{muppidi:2021:qcnn} (Geometric) & ConvNet (QCNN) & Emotion Classification & IEMO-CAP \cite{busso:2008:iemocapdataset} & 67.7 Mb & Acc: $70.46\%$ & N.A. & N.A. \\

    Muppidi and Radfar\cite{muppidi:2021:qcnn} (Geometric) & ConvNet (QCNN) & Emotion Classification & EMO-DB \cite{burkhardt:2005:berlinemodataset} & 31.2 Mb & Acc: $88.78\%$ & N.A. & N.A. \\

    Neshat \textit{et al.} \cite{neshat:2022:forecastingenergy} (Classic) & ConvNet (IAOA-QCNN-BiLSTM) & Forecasting: Wind speed & IERSD, NOA & N.A. & Lesbos's MSE: $7.8e^{-3}$; Samothraki's MSE: $1.27e^{-2}$ & N.A. & Lesbos's MSE: $1.06e^{-2}$; Samothraki's MSE: $1.31e^{-2}$ \\
    \hline
  \end{tabular*}
\end{table*}

\subsection{Vision}

For mapping data to the quaternion domain, several methods are available in the literature; in the case of color images, the most common approach is to encode the red, green, and blue channels into the imaginary parts of the quaternion:
\begin{equation}
  \mathbf{q}=0+R\hat{i}+G\hat{j}+B\hat{k},
\end{equation}
see for example \cite{beijing:2019:fullyqcnn, beijing:2021:regionqcnn, grassucci:2021:qcnn, grassucci:2022:qsngan, grassucci:2022:starganv2, hongo:2020:qcnn, jin:2020:deformablegaborqcnn, muppidi:2021:qcnn, parcollet:2019:qcnn, sfikas:2021:qgan, yin:2019:qcnn, zhu:2018:qcnn}.

Another method, proposed by Gaudet and Maida \cite{gaudet:2018:qcnn}, is to use a residual block:

$BN\rightarrow ReLU\rightarrow Conv\rightarrow BN\rightarrow ReLU\rightarrow Conv$,
where a shortcut connection sums the input to the output of the block.

In contrast, for grayscale images, Beijing \textit{et al.} \cite{beijing:2021:regionqcnn} propose to map the grayscale values to the real part of the quaternion as follows: 
\begin{equation}
  \mathbf{q}=grayvalue+0\hat{i}+0\hat{j}+0\hat{k}
\end{equation}

In the case of POLSAR images containing scattering matrices:
\begin{equation}
  \begin{bmatrix} S_{HH} & S_{HV} \\
  S_{VH} & S_{VV} \end{bmatrix}
\end{equation}
Matsumoto \textit{et al.} \cite{matsumoto:2022:fullyrotationalqcnn} propose to compute a Pauli decomposition from the complex scattering matrix:
\begin{equation}
  a = \frac{S_{HH}+S_{VV}}{\sqrt{2}},
  b = \frac{S_{HH}-S_{VV}}{\sqrt{2}},
  c = \frac{S_{Hv}+S_{Vh}}{\sqrt{2}}, 
\end{equation}
and assign the square magnitude of the components $b$, $c$, and $a$ to the red, green, and blue channels, respectively, of an RGB image. Thereafter, the image is mapped to the quaternion domain as usual. Alternatively, Matsumoto \textit{et al.} \cite{matsumoto:2022:fullyrotationalqcnn} propose to transform the scattering matrix into 3D Stokes vectors normalized by their total power. Thereafter, each component is mapped to the imaginary parts of the quaternion.

Finally, for point clouds, Shen \textit{et al.} \cite{shen:2020:qcnn} propose to map the $(x,y,z)$ coordinates of each 3D point to a quaternion as follows:
\begin{equation}
  \mathbf{q}=0+x\hat{i}+y\hat{j}+z\hat{k}.
\end{equation}

\subsubsection{Vision under the classic convolution paradigm}

Residual Networks were one of the first models to be proposed in the QCNNs literature. Gaudet and Maida \cite{gaudet:2018:qcnn} tested deep and shallow quaternion-valued residual nets. For both cases, they use three stages; the shallow network contains 2, 1 and 1 residual blocks in each stage, while the deep network uses 10, 9 and 9 residual blocks in each stage, see Figure \ref{fig:resqcnn}a). These models were applied on image classification tasks using the CIFAR-10 and CIFAR-100 \cite{krizhevsky:2009:cifar} datasets, and the KITTI Road Estimation benchmark \cite{fritsch:2013:kittiroadestimationbenchmark}. Recently, Sfikas \textit{et al.} \cite{sfikas:2022:greekmanuscript} proposed a standard residual model for keyword spotting in handwritten manuscripts, see Figure \ref{fig:resqcnn}c). Their model is applied on the PIOP-DAS \cite{sfikas:2022:piopdasdataset} and the GRPOLY-DB \cite{gatos:2015:grpolydataset} datasets, containing digitized documents written in modern Greek.

A different type of model is ConvNets. The first proposal came from Yin \textit{et al.} \cite{yin:2019:qcnn}, whom implemented a basic QCNN model, see Figure \ref{fig:qconvnets}f). Since quaternion models extract $4$ times more features, they propose to use an attention module, whose purpose is to filter out redundant features. These models were applied to the image classification problem on the CIFAR-10 dataset \cite{krizhevsky:2009:cifar}. In addition, they propose a similar model for Double JPEG compression detection on the UCID dataset \cite{huang:2010:uciddataset}. Thereafter, Jin \textit{et al.} \cite{jin:2020:deformablegaborqcnn} propose to include a deformable layer, leading to a Deformable Quaternion Gabor CNN (DQG-CNN), see Figure \ref{fig:qconvnets}h). They apply it on a facial recognition task using the Oulu-CASIA \cite{taini:2008:oulucasiadataset}, MMI \cite{pantic:2005:mmi:dataset}, and SFEW \cite{dhall:2011:sfewdataset} datasets. Beijing \textit{et al.} \cite{beijing:2019:fullyqcnn} propose an architecture based on quaternion versions of Fully Convolutional Neural Networks (FCN) \cite{shelhamer:2017:fcnn}. Their model is applied for the color image splicing localization problem, and tested on CASIA v1.0, CASIA v2.0 \cite{dong:2013:casiadataset}, and Columbia color DVMM \cite{ng:2004:dvmmdataset} datasets. The basis of their architecture is the Quaternion-valued Fully Convolutional Network (QFCN), see Figure \ref{fig:qconvnets}d). Then, the final model is composed of three QFCNs working in parallel, each one has different up-sampling layers: The first network does not have extra connections, the second network has a shortcut connection fusing the results of the fifth pooling layer with the output of the last layer, while the third network combines results of the third, and fourth pooling layers with the output of the last layer. In addition, each network uses a Super-pixel-enhanced pairwise Condition Random Field module to improve the results from the QCNN. This fusion of the outputs allows to work with different scales of image contents \cite{liu:2018:fcnn}. Thereafter, Beijing \textit{et al.} \cite{beijing:2021:regionqcnn} propose the quaternion-valued two-stream Region-CNN. They extend the real-valued RGB-N \cite{zhou:2018:regioncnn} to the quaternion domain, improve it for pixel-level processing, and implement two extra-modules: an Attention Region Proposal Network, based on CBAM \cite{woo:2018:cbam}, for enhancing the features of important regions; and a Feature Pyramid Network, based on Quaternion-valued ResNet, to extract multi-scale features. Their results improved the ones previously published on \cite{beijing:2019:fullyqcnn}. Because of the complexity of these models, it is recommended to read it directly from \cite{beijing:2019:fullyqcnn, beijing:2021:regionqcnn}. 

In the case of CAE's, which aim is to reconstruct the input feature at the output, Parcollet \textit{et al.} \cite{parcollet:2019:qcnn} propose a quaternion convolutional encoder-decoder (QCAE), see Figure \ref{fig:qgen}a), and tested on the KODAK PhotoCD dataset \cite{franzen:2013:kodakdataset}. 

Another type of CAE's is Variational Autoencoder, which estimates the probabilistic relationship between input and latent spaces. The quaternion-valued version was proposed by Grassucci \textit{et al.} \cite{grassucci:2021:qcnn}, see Figure \ref{fig:qgen}b), and applied it to reconstruct and generate faces of the CelebFaces Attributes Dataset (CelebA) \cite{liu:2015:celebfacesdataset}.

A more sophisticated type of architecture is the generative models. Sfikas \textit{et al.} \cite{sfikas:2021:qgan} propose a Quaternion Generative Adversarial Network for text detection of inscriptions found on byzantine monuments \cite{kordatos:2013:bizantinedataset, rhoby:2017:bizantinedataset}. The generator is a U-Net \cite{ronneberger:2015:unet} like model, and the discriminator a cascade of convolution layers, see Figures \ref{fig:qgen}e) and \ref{fig:qgen}f), where the activation function of the last layer sums the output of real and imaginary parts of the quaternion to produce a real-valued output.

Grassucci \textit{et al.} \cite{grassucci:2022:qsngan} adapted a Spectral Normalized GAN's (SNGAN) \cite{chen:2019:sngan, miyato:2018:sngan} to the quaternion domain, and apply it on an image to image translation task using the CelebA-HQ \cite{karras:2018:celebfaceshqdataset} and 102 Oxford Flowers \cite{nilsback:2008:oxfordflowers} datasets. The model is presented in Figure \ref{fig:qgen}c) and \ref{fig:qgen}d). Thereafter, Grassucci \textit{et al.} \cite{grassucci:2022:starganv2} proposed the quaternion-valued version of the StarGANv2 model \cite{choi:2020:starganv2}. It is composed of the generator, mapping, encoding, and discriminator networks; this model was evaluated on an image to image translation task using the CelebA-HQ dataset \cite{karras:2018:celebfaceshqdataset}. Because of the complexity of the model, it is recommended to consult them directly from \cite{grassucci:2022:starganv2}.

\subsubsection{Vision under the geometric paradigm}

The only residual model lying in this paradigm is the one of Hongo \textit{et al.} \cite{hongo:2020:qcnn}, whom proposed a residual model, based on ResNet34 \cite{he:2016:resnet39}, see Figure \ref{fig:resqcnn}a). This is applied to the image classification task using the CIFAR-10 dataset \cite{krizhevsky:2009:cifar}.

On ConvNet models, we have the work of Zhu \textit{et al.} \cite{zhu:2018:qcnn}, whom proposed shallow QCNN and quaternion VGG-S \cite{sermanet:2014:vggs} models for image classification problems. These models are shown in Figures \ref{fig:qconvnets}b) and \ref{fig:qconvnets}c); the former was tested on the CIFAR-10 dataset \cite{krizhevsky:2009:cifar}, and the latter on the 102 Oxford flower dataset \cite{nilsback:2008:oxfordflowers}. Hongo \textit{et al.} \cite{hongo:2020:qcnn} proposed a different QCNN model, see Figure \ref{fig:qconvnets}g), and tested on the image classification task using the CIFAR-10 dataset \cite{krizhevsky:2009:cifar}. Moreover, Matsumoto \textit{et al.} \cite{matsumoto:2022:fullyrotationalqcnn} propose QCNN models for classifying pixels of PolSAR images. This type of images contain additional experimental features given in complex scattering matrices. For their experiments, they labeled each pixel of two images in one of 4 classes: water, grass, forest, or town. Then, they converted the complex scattering matrices into PolSAR pseudocolor features, or into normalized Stoke vectors. By testing similar models under these two different representations, they found out that the classification results largely depend on input features. The proposed model is shown in Figure \ref{fig:qconvnets}j).

In the context of CAE's, Zhu \textit{et al.} \cite{zhu:2018:qcnn} propose a U-Net-like encoder-decoder network \cite{ronneberger:2015:unet} for the color image denoising problem. The model was tested for a denoising task on images of the 102 Oxford flower dataset \cite{nilsback:2008:oxfordflowers}, and on a subset of the COCO dataset \cite{tsung:2014:mscocodataset}, see details of the model in \cite{mao:2016:cae, ronneberger:2015:unet, vincent:2008:cae, zhu:2018:qcnn}.

\subsubsection{Vision under the equivariant paradigm}

Inspired by the PointNet model \cite{guibas:2017:pointnet} for processing point clouds, Shen \textit{et al.} \cite{shen:2020:qcnn} modified it by exchanging all its layers for Rotation Equivariant Quaternion Modules, and remove its Spatial Transformer module since it discards rotation information. The rotation equivariant properties of the modules are evaluated on the ShapeNet dataset \cite{chang:2015:shapenetdataset}. They experimentally proved that point clouds reconstructed using the synthesized quaternion features had the same orientations as point clouds generated by directly rotating the original point cloud. In addition, they modify other models, i.e. PointNet++ \cite{guibas:2017:pointnetpp}, DGCNN \cite{wang:2019:dgcnn}, and PointConv \cite{wu:2019:pointconv}, by replacing their components into its equivariance counterparts, and tested on a 3D shape classification task on the ModelNet40 \cite{wu:2015:3dshapenetsdataset} and 3D MNIST \cite{kaggke:2016:3dmnist} datasets. Because the variety of components and interconnections, the reader is directed to \cite{guibas:2017:pointnet, guibas:2017:pointnetpp, shen:2020:qcnn, wang:2019:dgcnn, wu:2019:pointconv} to consult the details of these models.

\subsection{Language}

For language applications, all works that have been proposed are ConvNets.

Under the classic paradigm, Parcollet \textit{et al.} \cite{parcollet:2018:qcnn} propose a Connectionist Temporal Classification CTC-QCNN model, see Figure \ref{fig:qconvnets}a), and tested on a phoneme recognition task, with TIMIT dataset \cite{1993:nist:timitdataset}. The mapping of input signals to the quaternion domain is achieved by transforming the raw audio into 40-dimensional log mel-filter-bank coefficients with deltas, delta-deltas, and energy terms, and arranging the resulting vector into the components of an acoustic quaternion signal:
\begin{equation}
  \mathbf{q(f,t)}=0+e(f,t)\hat{i}+ \frac{\partial e(f,t)}{\partial t} \hat{j} + \frac{\partial^2 e(f,t)}{\partial t^2} \hat{k}.
\end{equation}

Another problem related to language is joint 3D Sound Event Localization and Detection (SELD); solving this task can be helpful for activity recognition, assisting hearing impaired people, among other applications. The SELD problem consists in simultaneous solving: the Sound Event Detection (SED) problem, i.e. in a set of overlapping sound events, detecting temporal activities of each sound event and associating a label, and the Sound Localization Problem, which consists of estimating the spatial localization trajectory of each sound. The latter task could be simplified to determining the orientation of a sound source with respect to the microphone, which is called Direction-of-Arrival (DOA). For this problem, also following a classical approach, Comminiello \textit{et al.} \cite{comminiello:2019:qcnn} propose a quaternion-valued recurrent network (QSELD-net), based on the real-valued model of Adavanne \textit{et al.} \cite{adavanne:2019:seldnet}. The model has a first processing stage, where output is processed in parallel by two branches: the first one performs a multi-label classification task (SED), while the second one performs a multi-output regression task (DOA estimation). The model is shown in Figure \ref{fig:qconvnets}e), where $N$ is the number of sound event classes to be detected; for DOA estimation, we have three times more outputs, since they represent $(x,y,z)$ coordinates for each sound event class. This model was tested on the Ambisonic, Anechoic and Synthetic Impulse Response (ANSYN), and the Ambisonic, Reverberant and Synthetic Impulse Response (RESYN) datasets \cite{adavanne:2018:seldnet}, consisting of spatially located sound events in an anechoic/reverberant environment synthesized using artificial impulse responses. Each dataset comprises three subsets: no temporally overlapping sources (O1), maximum two temporally overlapping sources (O2) and maximum three temporally overlapping sources (O3). In this application, the input data is a multichannel audio signal; real-valued SELD-net \cite{adavanne:2019:seldnet} as well as the quaternion-valued counterpart \cite{comminiello:2019:qcnn} apply the same feature extraction method: the spectrogram is computed using the $M$-point discrete Fourier transform to obtained a feature sequence of $T$ frames containing the magnitude and phase components for each channel.

Under the geometric paradigm, Muppidi and Radfar \cite{muppidi:2021:qcnn} proposed a QCNN for the emotion classification from speech task and evaluate it on the RAVDESS \cite{livingstone:2018:ravnessdataset}, IEMOCAP \cite{busso:2008:iemocapdataset}, and EMO-DB \cite{burkhardt:2005:berlinemodataset} datasets. They convert speech waveform inputs from the time domain to the frequency domain using Fourier transform, thereafter compute its Mel-Spectrogram, and convert it to an RGB image. Finally, RGB images are processed with the model shown in Figure \ref{fig:qconvnets}i), where the Reset-ReLU activation function resets invalid values to the nearest point in color space.

\subsection{Forecasting}

Neshat \textit{et al.} \cite{neshat:2022:forecastingenergy} proposed an hybrid forecasting model, composed of QCNNs and Bi-directional LSTM recurrent networks, for prediction of short-term and long-term wind speeds. Historical meteorological wind data was collected from Lesvos and Samothraki Greek islands located in the North Aegean Sea, and obtained from the Institute of Environmental Research and Sustainable Development (IERSD), and the National Observatory of Athens (NOA). This work integrates classic standard QCNNs within a complete system, and shows that better results are obtained when using QCNNs than other AI models or handcrafted techniques. Beacuse of the complexity of the model, it is recommended to consult the details directly from \cite{neshat:2022:forecastingenergy}.

\subsection{Insights}\label{subsec:insights}

A fair comparison between the performance of the models and their individual blocks is difficult because of the fact that most of them are applied on different problems or using different datasets; moreover, most of the quaternion-valued models are based on their real-valued counterparts, and there are very few works based on incremental improvement over previous quaternion models. However, from the works reviewed, the following insights can be established:

\begin{enumerate}
  \item Quaternion-valued models achieve, at least, comparable results to their real-valued counterparts, and with a lower number of parameters. As was stated by \cite{sfikas:2022:greekmanuscript}: ``$\dots$an operation that is written as $y=Wx$, where $y\in\mathbb{R}^{4K}$, $x \in \mathbb{R}^{4L}$, and $W \in \mathbb{R}^{4K\times4L}$ is thus mapped to a quaternionic operation $\mathbf{y} = \mathbf{W} \mathbf{x}$, where $\mathbf{y} \in \mathbb{H}^K$, $\mathbf{x} \in \mathbb{H}^L$ and $\mathbf{W}\in H^{K\times L}$. Parameter vector $\mathbf{W}$ only contains $4 \times K \times L$ parameters compared to $4 \times 4 \times K \times L$ of $W$, hence we have a $4$x saving." In addition, some works report faster convergence on the training stage.
  \item For the image classification task, the CIFAR-10 dataset has become the standard benchmark. Although the classification errors reported in different works are not standardized, for example: some authors use data-augmentation techniques, the best reported result is from Gaudet and Maida \cite{gaudet:2018:qcnn}, whom uses a classic approach and a deep residual model. In addition, residual models present the best results for image classification tasks.
  \item Attention modules filter out redundant features, and their use can improve the performance of the network \cite{yin:2019:qcnn}.
  \item Variance Quaternion Batch Normalization (VQBN) experimentally outperforms Split Batch Normalization \cite{yin:2019:qcnn}.
  \item The integration of quaternion Gabor filters and convolution layers enhances the abilities of the model to capture features in the aspects of spatial localization, orientation selectivity, and spatial frequency selectivity \cite{jin:2020:deformablegaborqcnn, moya:2020:localphaseqcnn, moya:2021:monogenicqcnn}. 
  \item Quaternion generative models demonstrate better abilities to learn the properties of the color space, and generate RGB images with more defined objects than real-valued models \cite{grassucci:2022:qsngan, parcollet:2019:qcnn}.
\end{enumerate}

Finally, Table~\ref{tab:code} presents information about the available source code. It only includes source code published by the authors of the papers; other implementations can be found in the popular website \textit{paperswithcode} \cite{paperswithcode}.

\begin{table*}
  \caption{Available source code of some QCNNs ; all implementations lie in the classic approach.}
  \label{tab:code}
  \begin{tabular*}{\textwidth}{
    @{} p{0.275\textwidth} p{0.225\textwidth} p{0.45\textwidth} @{}}
    \hline
    Authors & Models & Website\\
    \hline
    Gaudet and Maida \cite{gaudet:2018:qcnn} & Residual & \url{https://github.com/gaudetcj/DeepQuaternionNetworks/} \\

    Grassuccii et al. \cite{grassucci:2021:qcnn, grassucci:2021:hypercomplexparam, grassucci:2022:starganv2, grassucci:2022:qsngan} & VAE, GAN, Hypercomplex & \url{https://github.com/eleGAN23} \\

    Parcollet \textit{et al.} \cite{parcollet:2018:qcnn,parcollet:2019:qcnn} & QRNNs, QLSTMs, QCNNs (ConvNets), QCAEs. & \url{https://github.com/Orkis-Research/Pytorch-Quaternion-Neural-Networks}\\

    Sfikas \textit{et al.} \cite{sfikas:2021:qgan} & GAN & \url{https://github.com/sfikas/quaternion-gan} \\
    \hline            
  \end{tabular*}
\end{table*}

\section{Further discussion}\label{sec:future}

Section \ref{sec:qcnnblocks} presented each individual block, discussion of current issues, knowledge gaps, and future works for improvement. Here, we discuss future directions of research and open questions for any model of QCNNs, independently of its components. Even more, some topics involve any type of quaternion-valued deep learning model.

\subsection{On proper comparisons}

When looking at the different models that have been proposed, one of the questions that arises is: Is it fair to compare a real-valued architecture with a quaternion-valued  one using the same topology and number of parameters? Note that in the first case, the optimization of the parameters occurs in an Euclidean $4$-dimensional space, while in the second case, the optimization occurs in the quaternion space, which can be connected to geometric spaces different to the Euclidean one. From this observation, we could argue that a more reasonable way to compare is: an optimal real value network vs the optimal quaternion-value network, even though they have different topologies, connections, and number of weights. Now, it is clear that a single real-valued convolution layer cannot capture interchannel relationships, but could a real-valued multilayer or recurrent architecture capture interchannel relationships without the use of Hamilton products? When are Hamilton products \textit{really} needed?

Something that could shed light on these questions is the reflection of Sfikas \textit{et al.} \cite{sfikas:2022:greekmanuscript}, whom believe that the effectiveness of quaternion networks is because of ``navigation on a much compact parameter space during learning" \cite{sfikas:2022:greekmanuscript}, as well as an extra parameter sharing trait.

\subsection{Mapping from data to quaternions}

Another question that arises is: how to find an optimal mapping from the input data to the quaternion domain? Common sense would say that finding a suitable mapping, and connecting with the right geometry could seriously improve the performance of quaternion networks; however, theoretical and experimental work is required in this direction. For example, for images, we can adopt different color models with a direct geometric relation to quaternion space, and point clouds as well as deformable models can avoid algebraic singularities when the quaternion representation is used. Moreover, models for language tasks are in their infancy, and novel methods that take advantage of the quaternion space should be proposed for signal processing, speech recognition, and other tasks.

\subsection{Extension to hypercomplex and geometric algebras}

Current QCNNs rely on processing 4-dimensional input data, which is mapped to the quaternion domain. If the input data have less than $4$ channels, a common solution is to apply zero-padding on some channels, or define a mapping to a $4$-dimensional space. 

In contrast, for more than $4$ channels, there are several alternatives: 

The first one is to map the input data to an $n$-dimensional space, where $n$ is a multiple of four. Then, we can process the input data by defining a quaternion kernel for each 4-channel input, see Figure \ref{fig:qconvschemes}. Alternatively, it can be applied an extension of the quaternion convolution to the octonions, sedenions or generalized hypercomplex numbers \cite{vieira:2022:hypercomplexcnn}. Moreover, Geometric algebras \cite{hestenes:1987:ca2gc, kenichi:2015:geometricalgebra} can be used to generalize hypercomplex convolution to general $n$-dimensional spaces, not restricted to multiples of $4$, and to connect with different geometries.

A partial approach in this direction is to parametrize the hypercomplex convolution \cite{grassucci:2021:hypercomplexparam, zhang2021:hypercomplexparam}. Let $W$ be a convolution kernel, of size $k\times k$, $x$ a $s$-dimensional input, and $y$ a $d$-dimensional output, then:
\begin{equation}
  y=W \ast x.
\end{equation}
Thus, the kernel is decomposed into the sum of its Kronecker products:
\begin{equation}
  y=\sum_{i=1}^{n}A_i \otimes F_i \ast x.
\end{equation}
where $A_i\in\mathbb{R}^{n\times n}$, with $i=1,\dots,n$ are the matrices containing the algebra multiplication rules, and $F_i\in\mathbb{R}^{\frac{s}{n}\times \frac{d}{n}\times k\times k}$ are the filters that compose the final weight matrix. The parameter $n$ defines the dimension of the algebra; then, for $n=2$ we are working in the complex domain, while a value $n=4$ leads us to the quaternion space. Matrices $A_i$, and $F_i$ are obtained during training.

\subsection{Quaternions on the frequency domain}

A natural implementation of QCNNs in the frequency domain would be to compute the Quaternion Fourier transform \cite{bulow:2001:quaternionpolar, hitzer:2016:qft_properties, pei:2013:qft_fast} of the input and kernels, and multiply them in the frequency domain. Moreover, the computing of the convolution could be accelerated using Fast QFT algorithms \cite{said:2008:qfft}. To this date, works following this approach have not been published. However, using a bio-inspired approach, Moya-sanchez \textit{et al.} \cite{moya:2021:monogenicqcnn} proposed a monogenic convolution layer. A monogenic signal \cite{felsberg:2001:monogenicsignal}, $I_M$, is a mapping of the input signal, which simultaneously encodes local space and frequency characteristics:
\begin{equation}
  I_M = I'+I_1\hat{i}+I_2\hat{j}
\end{equation}
where $I_1$ and $I_2$ are the Riesz transforms of the input in the $x$ and $y$ directions. In \cite{moya:2021:monogenicqcnn}, the monogenic signal is computed in the Fourier domain, $I'$ is computed as a convolution of the input with quaternion log-Gabor filters with learnable parameters, and local phase and orientation are computed for achieving contrast invariance. In addition, their model provides crucial sensitivity to orientations, resembling properties of V1 cortex layer.

\subsection{Might the classic, geometric, and equivariant models be special cases of a unified general model?}

To answer this question, let us recall that a quaternion is an element of a 4-dimensional space, and unitary versors together with the Hamilton product are isomorphic to the group SO(4). Thus, it is our point of view that the classic model, using the four components, is the most general one. By dividing the product as a sandwiching product and using an equivalent polar representation, a generalized version of the geometric model could be obtained (not Euclidean or affine, but 3D projective model); in addition, for the equivariant model, the kernel is reduced merely to its real components. However, the connection to invariant theory should be investigated for each of these models, so that a stratified organization at the light of geometry might be achieved. For example, if we link a unified general model to a particular geometry and its invariant properties, we could go down on the hierarchy of geometries by setting restrictions on the general model, until we obtain the rotation equivariant model or the others.

\section{Conclusions}\label{sec:conclusions}

We have recollected the substantial majority of the ideas that have been published on QCNNs in recent years, and we presented a comprehensive guide for its application. In particular, being the convolution layer the core component of this type of models, we presented a sounded organization of QCNNs based on the definition of convolution; therefore, we proposed three classifications: classic, geometric, and rotation equivariant. For other components, we presented the purpose of each block, the key problem to solve in its design, knowledge gaps, if any, and ideas for future improvement. In addition, a review of the models by application domain and topology type was presented, including available source code. Further ideas for model design were also discussed.

Finally, most of the ideas that have been implemented are extensions of the work on real-valued CNNs or complex CNNs to the quaternion domain \cite{guberman:2016:c2n2, trabelsi:2018:complexnn}, or adaptations from quaternion neural networks models \cite{parcollet:2020:qnnsurvey}. Further work is required in developing novel ideas, and exploiting the particularities of quaternion representation and its connection to geometry, topology, functional analysis, or invariant theory. This paper presents, in an organized manner, the current advances in the development of QCNNs, in the hope that it serves as a starting point for subsequent research, as well as for those interested in implementing applications.

\section*{Acknowledgments}

G.A.G. received a Postdoctoral Fellowship \textit{Estancias Posdoctorales por México} from CONACYT. C.G. acknowledges support from UNAM-PAPIIT (IN107919, IV100120, IN105122) and from the PASPA program from UNAM-DGAPA.

\bibliographystyle{IEEEtran}
\bibliography{qcnn}

\begin{IEEEbiographynophoto}{Gerardo Altamirano-Gomez} holds a B.E. in Mechatronics from the National Polytechnic Institute (IPN, México), a M.S. in Computer Science from the  Center for Scientific Research and Higher Education at Ensenada (CICESE, México), and a D.Sc. in Electric Engineering from the Center for Research and Advanced Studies of the National Polytechnic Institute (CINVESTAV-IPN, México). He has lectured in several universities across Mexico, and is co-founder of the Instituto de Cómputo Aplicado en Ciencias S.C., México. Since 2021, he has been a Postdoctoral Researcher at Universidad Nacional Autónoma de México (UNAM, México). His research interests lie in the intersection of computer vision, artificial intelligence and geometric algebra computing. \end{IEEEbiographynophoto}

\begin{IEEEbiographynophoto}{Carlos Gershenson} is a tenured, full time research professor at the computer science department of the Instituto de Investigaciones en Matemáticas Aplicadas y en Sistemas at the Universidad Nacional Autónoma de México (UNAM, Mexico), where he leads the Self-organizing Systems Lab. He is also an affiliated researcher at the Center for Complexity Sciences at UNAM, where he was coordinator of the Complexity Sciences National Laboratory (2017-2022). He is currently a Visiting Researcher at the Santa Fe Institute. He was a Visiting Professor at MIT and at Northeastern University (2015-2016) and at ITMO University (2015-2019). He was a postdoctoral fellow at the New England Complex Systems Institute (2007-2008). He holds a PhD  summa cum laude from the Vrije Universiteit Brussel, Belgium. He holds an MSc degree in Evolutionary and Adaptive Systems, from the University of Sussex, and a BEng degree in Computer Engineering from the Fundación Arturo Rosenblueth, México. He studied five semesters of Philosophy at UNAM (1998-2001). He has been an active researcher since 1997, working at the Chemistry Institute, UNAM, México, and a summer (1999) at the Weizmann Institute of Science, Israel. He has more than 150 scientific publications in books, journals, and conference proceedings, which have been cited more than 6300 times. He has given more than 300 presentations at conferences and research group seminars. He is Editor-in-Chief of Complexity Digest, Associate Editor for the journal Complexity, and member of the Board of Advisors for Scientific American. He has received numerous awards, including a Google Research Award in Latin America and the Audi Urban Future Award. He is a member of the Mexican Academy of Sciences and the Mexican Academy of Informatics.
\end{IEEEbiographynophoto}

\end{document}